%% file: arxiv.tex
\documentclass[11pt, a4paper, logo, copyright, nonumbering]{style}
\usepackage[authoryear, sort&compress, round]{natbib}
\usepackage{graphicx}
\usepackage{xspace}
\usepackage{xcolor}
\usepackage{adjustbox}
\usepackage{array}
\usepackage{float}
\usepackage{geometry}
\usepackage{dblfloatfix}
\usepackage{enumitem}
\usepackage{setspace}
\usepackage{footmisc}
\usepackage{booktabs}
\usepackage{multirow}
\usepackage{longtable}
\usepackage{tabularx}
\usepackage{xltabular}
\usepackage{threeparttable}
\usepackage{siunitx}
\usepackage{mathtools,amsfonts,amsthm,amssymb,bbm,amsmath}

\usepackage[toc,page,header]{appendix}
\usepackage{minitoc}
\usepackage{wrapfig}

\usepackage{subcaption} 

\usepackage[utf8]{inputenc} 
\usepackage[T1]{fontenc}    
\usepackage{hyperref}       
\usepackage{url}            
\usepackage{booktabs}       
\usepackage{amsfonts}       
\usepackage{nicefrac}       
\usepackage{microtype}      
\usepackage{xcolor}         
\usepackage{graphicx}       
\usepackage{amsmath}        
\usepackage{amssymb}
\usepackage{multirow}
\usepackage{amsfonts}
\usepackage{mathtools}
\usepackage{bm}
\usepackage{dsfont}

\usepackage{subcaption}
\usepackage{tikz}
\usepackage{pgfplots}
\pgfplotsset{compat=1.18}
\usepackage{algorithm}
\usepackage{algpseudocode}

\usepackage{cleveref}
\usepackage{hyperref}
\hypersetup{colorlinks=true}

\usepackage{todonotes}
\usepackage{fontawesome5}
\usepackage[most]{tcolorbox}
\usepackage{CJKutf8}
\usepackage{lipsum}
\usepackage{amsthm}
\theoremstyle{plain}
\newtheorem{proposition}{Proposition}
\theoremstyle{definition}
\newtheorem{definition}{Definition}
\newtheorem{remark}{Remark}
\input{macros}

\makeatletter
\def\@BTrule[#1]{%
  \ifx\longtable\undefined
    \let\@BTswitch\@BTnormal
  \else\ifx\hline\LT@hline
    \nobreak
    \let\@BTswitch\@BLTrule
  \else
     \let\@BTswitch\@BTnormal
  \fi\fi
  \global\@thisrulewidth=#1\relax
  \ifnum\@thisruleclass=\tw@\vskip\@aboverulesep\else
  \ifnum\@lastruleclass=\z@\vskip\@aboverulesep\else
  \ifnum\@lastruleclass=\@ne\vskip\doublerulesep\fi\fi\fi
  \@BTswitch}
\makeatother

\addto\extrasenglish{
}

 {\begin{list}{}%
         {\setlength{\leftmargin}{#1}}%
         \item[]%
 }
 {\end{list}}
 
\title{\centering A Simple "Try Again" Can Elicit Multi-turn LLM Reasoning}

\author{
Licheng Liu\textsuperscript{1*}, 
Zihan Wang\textsuperscript{2*}, 
Linjie Li\textsuperscript{3}, 
Chenwei Xu\textsuperscript{2},
Yiping Lu\textsuperscript{2}, 
Han Liu\textsuperscript{2}, 
Avirup Sil\textsuperscript{4}, 
Manling Li\textsuperscript{2}
\\
\small \textsuperscript{1}Imperial College London~~ 
\textsuperscript{2}Northwestern University~~
\textsuperscript{3}University of Washington~~
\textsuperscript{4}IBM Research AI
\\
\small \href{https://unary-feedback.github.io}{\texttt{unary-feedback.github.io}}
}

\date{}

\input{commands.tex}
\input{sections/00_abstract}

\newcounter{examplebox}

\begin{document}

\doparttoc 
\faketableofcontents 
\maketitle

\input{sections/01_intro}
\input{sections/02_method}

\input{sections/03_experiments}

\input{sections/04_analysis}
\input{sections/05_related_work}
\input{sections/06_imp_ccl}

\medskip
{
\small
\bibliographystyle{unsrtnat}
\bibliography{main}
}
\clearpage
\renewcommand \thepart{}
\renewcommand \partname{}
\addcontentsline{toc}{section}{Appendix} 
\appendix
\part{Appendix} 
\parttoc 

\input{sections/07_appendix_supplementary}

\end{document}

%% file: macros.tex
\usepackage{xspace}

\definecolor{MyOrange}{rgb}{1.0, 0.4, 0.2}
\definecolor{MyBlue}{rgb}{0.2, 0.2, 0.8}

\renewcommand\paragraph[1]{\vspace{0em}\noindent\textbf{#1}}
\usepackage[font=footnotesize,labelfont=bf,aboveskip=2pt,belowskip=-13pt]{caption}
\usepackage[most]{tcolorbox}
\usepackage{xcolor}

\newcounter{promptcnt} 
\newcounter{casecnt}

\newtcblisting{prompt}[2][Prompt Template]{%
    enhanced,
    breakable,
    colback=gray!5,
    colframe=gray!40,
    listing only,
    listing options={
        basicstyle=\ttfamily\small,
        columns=flexible,
        breaklines=true,
        keepspaces=true,
        breakindent=0em,
        keepspaces=true,
        escapechar=\%,
        upquote=true
    },
    title={#1}, 
    attach boxed title to top left={xshift=0.5cm, yshift=-\tcboxedtitleheight/2},
    before upper={\refstepcounter{promptcnt} \label{#2}}, 
}

\newtcblisting{case}[2][Case Template]{%
    enhanced,
    breakable,
    colback=red!8,
    colframe=red!60,
    coltitle=white,
    colbacktitle=red!80!magenta!70,
    listing only,
    listing options={
        basicstyle=\ttfamily\small,
        columns=flexible,
        breaklines=true,
        keepspaces=true,
        breakindent=0em,
        keepspaces=true,
        escapechar=\%,
        upquote=true
    },
    title={#1}, 
    attach boxed title to top left={xshift=0.5cm, yshift=-\tcboxedtitleheight/2},
    before upper={\refstepcounter{casecnt} \label{#2}}, 
}
\usepackage{xcolor} 
\definecolor{mybluetitle}{rgb}{0.2,0.4,0.7} 
\newtcolorbox{onebox}[2][]{
    enhanced, 
    center title,
    left*=0pt, right*=0pt,
    boxsep=2pt, left=5pt, right=5pt,
    skin first=enhanced,
    skin middle=enhanced,
    skin last=enhanced,
    colframe = mybluetitle!90,
  colback  = mybluetitle!10,
    fonttitle=\bfseries\rmfamily\fontfamily{phv}\selectfont,
    title={\footnotesize\strut{#2}  \refstepcounter{subsubsection} \addcontentsline{toc}{subsubsection}{\string\numberline{\thesubsubsection}#2}
    },
    #1
    }

%% file: commands.tex
\renewcommand{\phi}{\varphi}

\renewcommand{\leq}{\leqslant}

\renewcommand{\epsilon}{\varepsilon}
\renewcommand{\imath}{\mathrm{i}}

\newlength{\restsubwidth}
\newlength{\restsubheight}
\newlength{\restsubmoreheight}
\newcommand{\rest}[2]{%
        \settowidth{\restsubwidth}{\ensuremath{#2}}
        \settoheight{\restsubheight}{\ensuremath{{}_{#2}}}
        \ensuremath{{#1\hskip 0.5pt}_{\vrule\kern2pt\parbox[b][%
        4pt][b]{\the\restsubwidth}{%
                        \ensuremath{{}_{#2}}}}}
        }

%% file: sections/00_abstract.tex
\begin{abstract}
    Multi-turn problem solving is critical yet challenging for Large Reasoning Models (LRMs) to reflect on their reasoning and revise from feedback. Existing Reinforcement Learning with Verifiable Reward (RLVR) methods train large reasoning models on a single-turn paradigm. However, we observe that models trained with existing RL paradigms often \textbf{fail to explore alternative reasoning paths across multiple turns} and lack the capacity for self-reflection, resulting in repetitive and unadapted responses to contextual feedback.
    We ask: Can LRMs learn to reflect their answers in a multi-turn context? In this work, we find that training models with multi-turn RL using only unary feedback (for example, “Let’s try again”) after wrong answers can improve both single-turn performance and multi-turn reasoning. We introduce \textbf{Unary Feedback as Observation (UFO)} for reinforcement learning, which uses minimal yet common unary user feedback during iterative problem solving. It can be easily applied to existing single-turn RL training setups. Experimental results show that RL training with UFO keeps single-turn performance and improves multi-turn reasoning accuracy by up to 14\%, enabling models to reflect on prior failures and refine their reasoning accordingly. To further minimize the number of turns needed for a correct answer while encouraging diverse reasoning when mistakes occur, we design reward structures that guide models to produce careful and deliberate answers in each turn. Our code and models are open source \href{https://github.com/lichengliu03/unary-feedback}{https://github.com/lichengliu03/unary-feedback}.
\end{abstract}

%% file: sections/01_intro.tex
\section{Introduction}

\begin{figure*}[t]
    \vspace{-10pt}
    \centering
    \includegraphics[width=\textwidth]{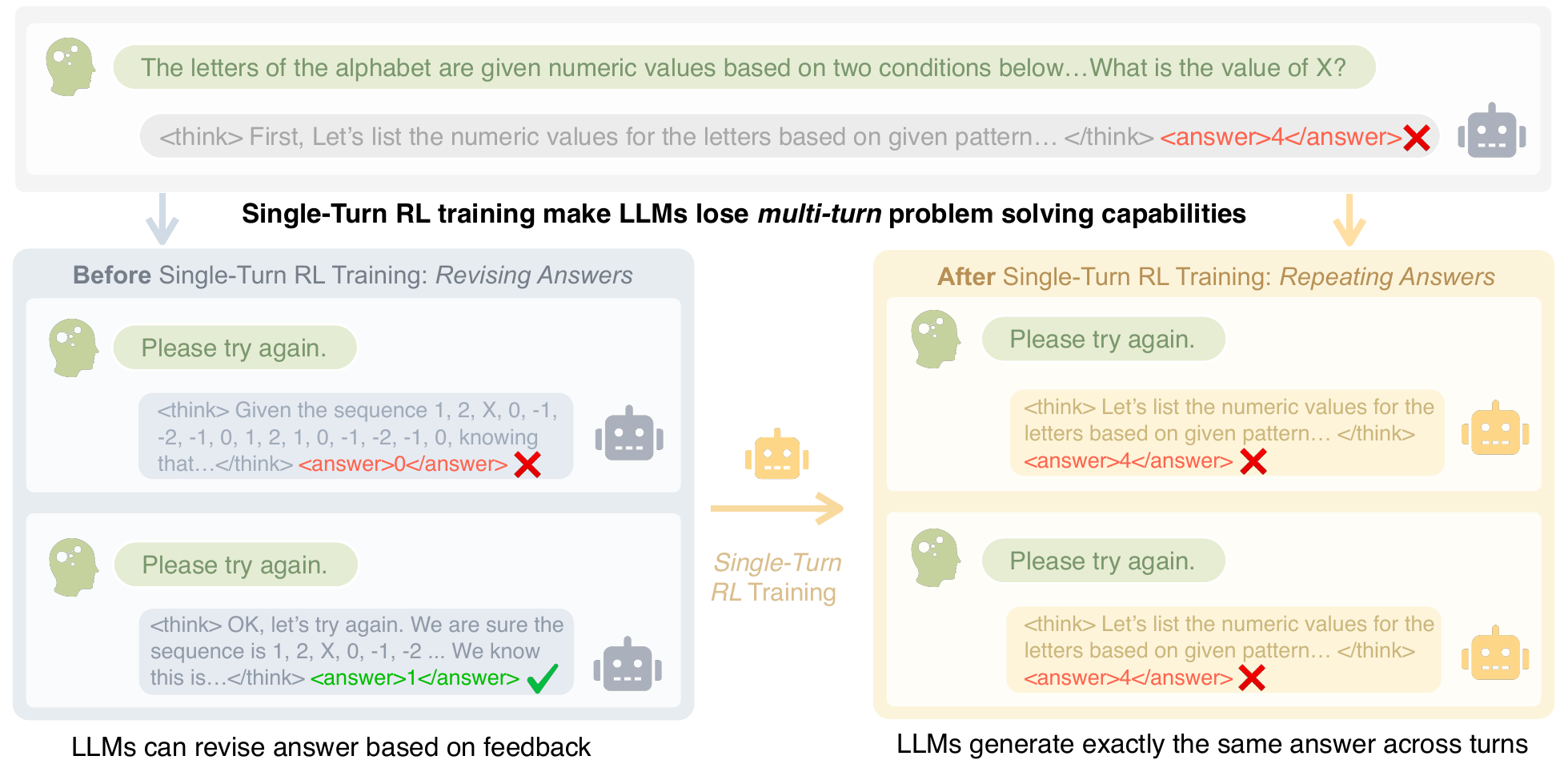}
    \vspace{5pt}
    \caption{Single-turn RL causes LLMs to repeat the same answer across turns instead of revising based on feedback.}
    \label{fig:example}
\end{figure*}

Large language and reasoning models (LLMs/LRMs)~\citep{deepseekr1, gpt4, qwen25, gemini} have demonstrated strong capabilities in solving complex tasks such as mathematics problems and code generation. Recent advances in reinforcement learning (RL)~\citep{deepseekr1, ppo, archer, ragen} further enhance the reasoning capabilities of LLMs through verifiable reward frameworks (RLVR). However, many real-world applications like chatbots, programming assistants, and educational tools~\citep{osworld, swegym, webshop, alfworld, mint} require models to engage in multi-turn problem solving and adapting their reasoning with feedback. Despite this need, it remains underexplored how models trained with single-turn RL can generalize to interactive, multi-turn problem-solving.

\begin{wrapfigure}{r}{0.42\textwidth}
    \vspace{-12pt}
    \centering
    \includegraphics[width=\linewidth]{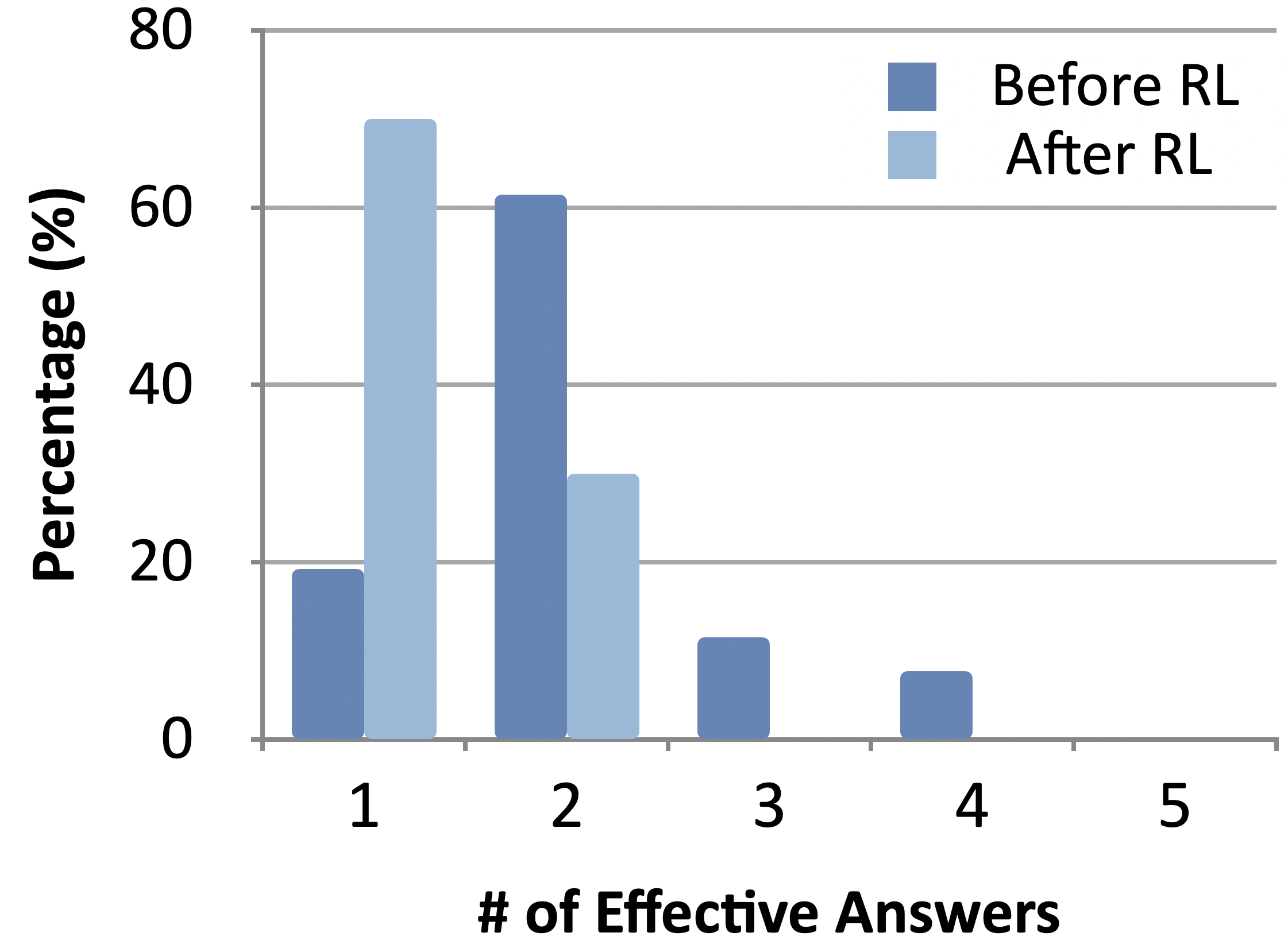}
    \vspace{-5pt}
    \caption{After single-turn RL training, the model gives exactly the same answer across five interaction turns in 70\% of the cases.}
    \label{fig:answer_repetition}
\end{wrapfigure}

In this work, we first observe that single-turn RL can hinder a model’s ability to engage in interactive multi-turn reasoning. Specifically, single-turn-trained models often fail to incorporate in-context feedback and instead \textbf{persist with their initial answers across subsequent turns} (Figure~\ref{fig:example}). To quantify this persistence, we define \textit{effective answer} as a new response that has never been explored in any previous turn by the LLM. In Figure~\ref{fig:answer_repetition}, we observed that in 70\% of the failed cases, the single-turn-trained models generate exactly the same answer across five turns (i.e. only 1 effective answer). One particular reason for that is that most existing datasets are inherently single-turn and lack signals for iterative exploration. Without such signals, models have limited opportunities to explore diverse reasoning paths or adapt their strategies based on feedback. This challenge motivates our research question: \textit{How can we train language models to iteratively explore and refine their reasoning, especially when initial attempts fail and only minimal feedback is available?}

Collecting real-world multi-turn user feedback, however, is expensive and logistically difficult. Constrained by this data bottleneck, the existing multi-turn framework has been training in automatic feedback such as code interpreter messages~\citep{osworld, swegym, mint} and embodied simulator signals~\citep{alfworld, simworld}. Although useful, these signals are costly to construct and are still limited in scope~\citep{cao2025skyrl}. In light of these limitations, we explore a simple yet effective framework that can leverage static dataset for multi-turn RL training. By inserting minimal verbal feedback ("try again") into the context, we encourage models to revise and explore alternative reasoning paths. We call this \textbf{Unary Feedback as Observation} (UFO), framing interactive problem-solving as Markov Decision Processes (MDP) where the model only receives unary feedback as its environmental observation.

Through experiments, we show that applying UFO in multi-turn RL effectively encourages interactive reasoning and helps the model explore alternative reasoning paths across turns. Instead of repeating previous answers, UFO trained models adjust their problem-solving strategies based on prior outcomes, leading to a 14\% improvement in multi-turn success rates over conventional single-turn RL. Furthermore, we find that the model’s ability to self-reflect and adapt its reasoning generalizes robustly across diverse out-of-domain tasks. 

To further align model behavior with real-world multi-turn reasoning objectives, we propose two guiding principles: \textbf{minimality}, where models should minimize the number of interaction turns needed to reach a correct answer, and \textbf{diversity}, where they should explore varied strategies when faced with failure. To operationalize these principles, we introduce a turn-wise reward decay and an answer repetition penalty which encourage systematic planning and improve reasoning efficiency.

To summarize, our contributions are as follows:
\begin{itemize}[leftmargin=*, itemsep=0em, topsep=0pt]
    \item We identify that while current single-turn RL training improves reasoning, they can lead to repetitive and degraded outputs in multi-turn, interactive reasoning scenarios.
    \item We explore a simple yet effective framework, \textbf{Unary Feedback as Observation (UFO)}, to enable multi-turn RL training on existing static single-turn reasoning datasets.
    \item We show that turn-wise reward decay and answer repetition penalty could effectively improve multi-turn reasoning minimality and diversity.
\end{itemize}

\begin{figure*}[t]
    \centering
    \vspace{-12pt}
    \includegraphics[width=0.98\linewidth]{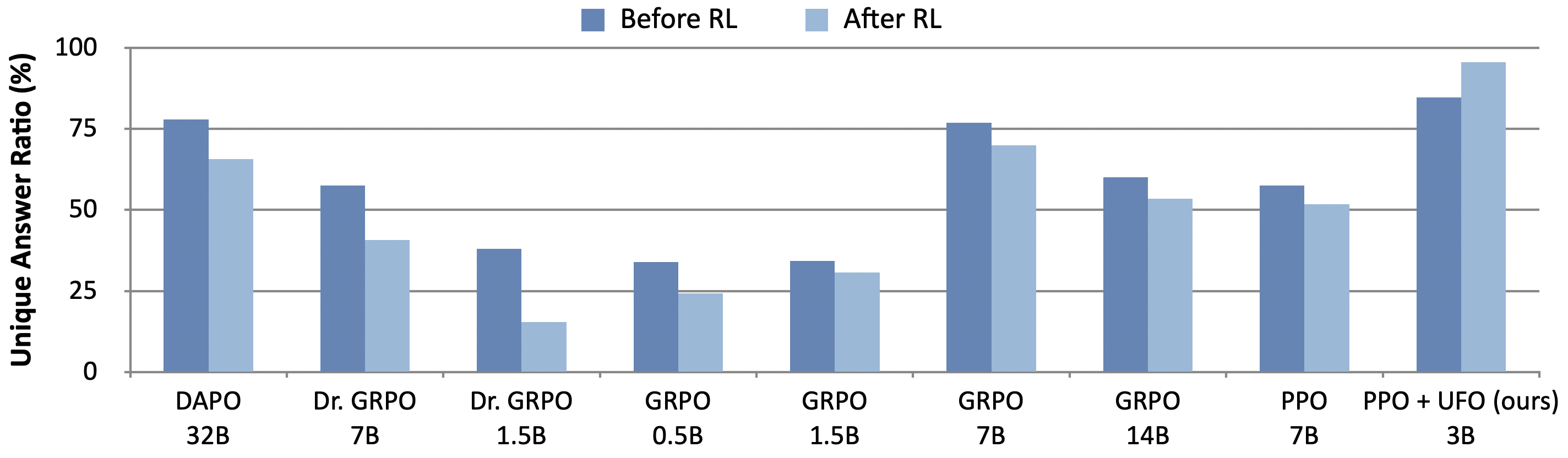}
    \vspace{10pt}
    \caption{\textbf{Comparison of effective (unique) answer ratio (\%) before and after RL training}. Across single-turn RL methods, the unique answer ratio consistently drops after training across multiple model scales.}
    \vspace{5pt}
    \label{fig:compare_unique}
\end{figure*}

%% file: sections/02_method.tex
\section{Reinforcement Learning for LLM Reasoning}

\subsection{Background}
\paragraph{Single-Turn Reinforcement Learning.}
Reinforcement Learning (RL) is a general framework to steer the behavior of LLMs by maximizing expected reward:
\begin{align*}
\mathbb{E}_{x \sim \mathcal{D},\, y \sim \pi_\theta(\cdot \mid x)}[R(x,y)],
\end{align*}
where $\mathcal{D}$ is a prompt distribution, $\pi_\theta$, is the LLM policy parameterized by $\theta$, and $R(x,y)$ is the reward for response $y$. Algorithms such as PPO~\citep{ppo,instructgpt} and GRPO~\citep{deepseekr1,deepseekmath} apply this objective to static datasets, yielding strong single-turn gains in math and code generation. 

\paragraph{Multi-Turn Extensions.}
Though these methods excel at optimizing a policy to produce a correct answer in a single attempt, real-world applications like tutoring, coding assistants, embodied agents demand \emph{multi-turn} interaction, where a model refines answers across steps under feedback. In domains like programming, automated feedback is readily available from compilers or unit tests, enabling iterative correction. But for tasks like mathematical reasoning, obtaining such dense feedback is difficult; often, only a final signal of correctness is available.

Recent work has begun addressing the challenge of sparse feedback in reasoning tasks by optimizing full interaction trajectories.
For example, CollabLLM~\citep{collabllm2025} introduces multi-turn fine-tuning with collaborative simulation and sparse rewards, while RAGEN~\citep{ragen} frames reasoning as an MDP to enable delayed credit assignment. However, these methods often depend on custom environments or simulated rewards, limiting scalability. Since most real-world math and code datasets are single-turn and collecting turn-level human feedback is costly, some efforts synthesize proxy signals~\citep{swegym, osworld} or build tool-augmented environments~\citep{searchr1, retool, mint}. Despite these workarounds, most training remains fundamentally grounded in single-turn RL paradigms.

These limitations give rise to a fundamental question:
\emph{Can models trained only with single-turn RL generalize to multi-turn reasoning?}

\begin{figure*}[t]
    \centering
    \includegraphics[width=0.98\linewidth]{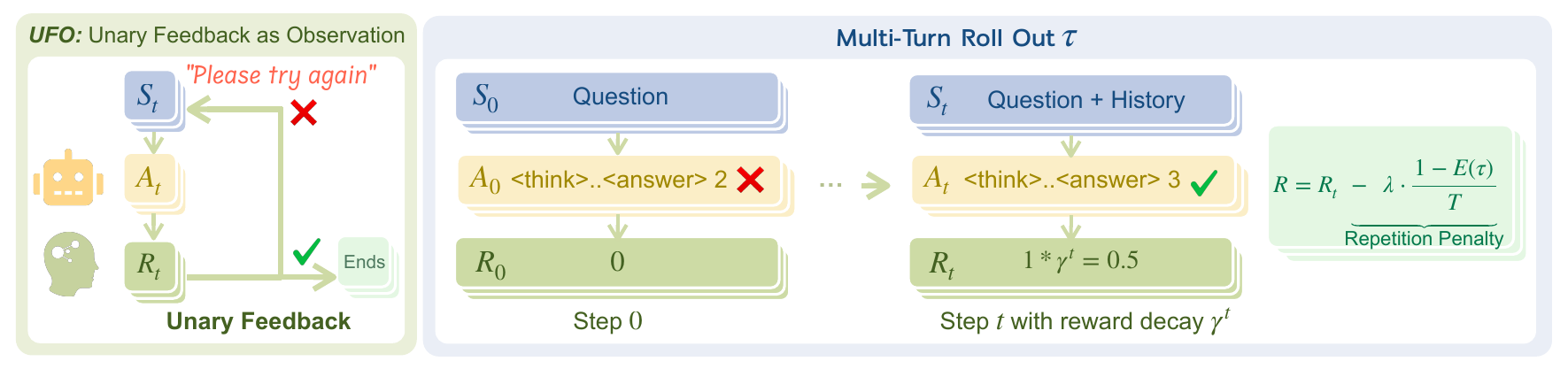}
    \vspace{10pt} 
    \caption{\textbf{The UFO framework for multi-turn training.} At each step $t$, the model observes the full interaction history and generates a response. Correct responses receive discounted rewards $\gamma^t$, while incorrect ones receive none. A repetition penalty based on the uniqueness of trajectory $\tau$ is applied after success or when the turn limit is reached.
    }
    \label{fig:ufo}
\end{figure*}

\subsection{Single-Turn RL Leads to Collapsed Multi-Turn Reasoning}

To answer the question, we need to examine how models trained with single-turn RL perform in multi-turn setting. Practically, users typically offer minimal feedback (e.g., “try again”) and expect the model to adjust its reasoning accordingly. However, we find that single-turn RL models are \textbf{effective solvers but poor revisers}, consistently failing to incorporate feedback.

This phenomenon is illustrated in Figure~\ref{fig:example}: a pre-trained model refines its answer across turns, while a single-turn RL model fails to revise, repeating its initial output. To quantify this behavior, we use \emph{effective answer} as our metric, and the results were presented in Figure~\ref{fig:compare_unique} for off-the-shelf LLMs and Figure~\ref{fig:answer_repetition} for our trained models with single-turn RL. Empirically, after single-turn RL training, LLMs tend to give less effective answers across multiple turns.

Specifically, for off-the-shelf LLMs, we select models fine-tuned with various RL algorithms including PPO~\citep{ppo}, GRPO~\citep{deepseekmath}, DAPO~\citep{yu2025dapoopensourcellmreinforcement}, and Dr.\ GRPO~\citep{liu2025understanding}. Details of the models used can be found in the Appendix~\ref{sec:model-detail}. As shown in Figure~\ref{fig:compare_unique}, all models show a noticeable decrease in the unique answer ratio after RL training, and the extent varies by method and model size.
For example, under DAPO the effective answer ratio of the 32B model falls from 78.0\% to 65.7\%, and under Dr.\ GRPO that of the 1.5B model drops from 38.0\% to just 15.4\%. On the contrary, GRPO shows more moderate losses (e.g., the 0.5B model decreased from 34.0\% to 24.3\%), and the impact of PPO is also mild (the 7B model went from 57.6\% to 51.7\%). 
We also measure how many distinct answers our single-turn RL model provides for questions it fails to answer correctly for 5 consecutive turns (see Section~\ref{sec:exp_setup} for details). As shown in Figure~\ref{fig:answer_repetition}, under nearly 70\% cases the model provide identical wrong answers across multiple interaction turns.

\subsection{Theoretical Analysis}
\label{sec:theoretical_sketch}

We theoretically analyze why single-turn RL models tend to repeat mistakes in multi-turn settings. This behavior is a predictable consequence of the training process itself, which creates a \textbf{peaked, low-entropy output distribution}~\citep{cui2025entropymechanismreinforcementlearning, yue2025doesreinforcementlearningreally}. To quantify this tendency, we use the \emph{collision probability}, the likelihood that two independent samples are identical, which is formally defined as:
\begin{equation}
    \mathrm{Coll}(q) := \sum_y q(y \mid x)^2.
\end{equation}
This probability is fundamentally lower-bounded by the distribution's Shannon Entropy, $\mathcal{H}(q)$, as captured by the inequality:
\begin{equation}
    \mathrm{Coll}(q) \ge \exp(-\mathcal{H}(q)).
\end{equation}
We formalize this relationship with greater rigor and provide detailed proofs in Appendix~\ref{sec:app-repetition-theory}. This relationship proves that as a model's policy becomes more deterministic (lower entropy), the probability of repetition is forced to increase. For a static policy unable to learn from feedback, repetition is therefore an inevitable failure mode.

To overcome this limitation, a policy must be able to learn from its interaction history. We formalize this requirement using a Markov Decision Process (MDP) and define two distinct classes of policies. \textbf{Parallel Policies ($\Pi_{par}$)} correspond to traditional single-turn RL, where $k$ answers are sampled independently based only on the initial question. In contrast, \textbf{Sequential Policies ($\Pi_{seq}$)} make decisions based on the full history of interactions, allowing for adaptation after each failure. Since any parallel policy is a special case of a sequential one, it is clear that $\Pi_{par} \subseteq \Pi_{seq}$, and their expected success rates provably satisfy:
\begin{equation}
    \max_{\pi \in \Pi_{seq}} \mathbb{E}[\text{Succ@}k] \ge \max_{\pi \in \Pi_{par}} \mathbb{E}[\text{Succ@}k].
\end{equation}
This theoretical guarantee stems from the fact that a sequential policy can intentionally avoid previously known errors, thus increasing its subsequent conditional success probability. By Blackwell dominance~\citep{Blackwell1951}, its overall success rate is guaranteed to be at least as high as that of a parallel policy. We extend more detailed analysis in Appendix~\ref{app:theory-ufo}.

This highlights a critical gap: single-turn RL is insufficient for multi-turn reasoning. However, acquiring the necessary step-by-step supervision is often infeasible. In light of this, we ask the following question: \textbf{Can we leverage only the simplest form of supervision, such as “try again”, to simulate multi-turn interaction on static datasets and train models to learn adaptive revision behaviors?}
\begin{quote}
    \emph{Can minimal feedback alone unlock multi-turn reasoning on static datasets?}
\end{quote}

\section{Training Multi-Turn Reasoning Models with Unary Feedback}
\subsection{Problem Formulation}

We model the process of multi-turn problem solving based on static single-turn datasets as a finite-horizon Markov Decision Process (MDP), defined by the tuple $(\mathcal{S}, \mathcal{A}, \mathcal{P}, R, T_{\text{max}})$. Here, $\mathcal{S}$ is the state space, $\mathcal{A}$ is the action space consisting of all possible answers, $\mathcal{P}$ is the transition function defined by the agent–environment interaction, $R$ is the reward function, and $T_{\text{max}}$ is the maximum number of interaction steps per episode.
At each turn $t$, the agent observes a state $s_t \in \mathcal{S}$ that encodes the original question $q$ and the history of past attempts and feedbacks:
\begin{align}
    s_t = \texttt{Concat}(q, \{(a_k, f_k)\}_{k=1}^{t-1}),
\end{align}
where $a_k$ denotes the $k$-th answer, and $f_k$ is a feedback token returned by the environment. The agent then generates an answer $a_t \sim \pi_\theta(\cdot \mid s_t)$ and receives a scalar reward:
\begin{align}
    r_t = 
    \begin{cases}
        1, & \text{if } a_t \text{ is correct}, \\
        0, & \text{otherwise}.
    \end{cases}
\end{align}
The episode ends when the agent provides a correct answer or reaches the maximum number of steps $T_{\text{max}}$.
This formulation grounds the multi-turn learning problem in a standard RL framework.

\subsection{Unary Feedback as Observation (UFO)}

To implement the MDP described above on static datasets, we propose a simple yet general mechanism called \textbf{Unary Feedback as Observation (UFO)} (Figure~\ref{fig:ufo}). This mechanism defines how the state history is constructed and presented to the agent. The key idea is to restrict $f_k$ in the observation to negative signals only. Specifically, when an answer $a_k$ is incorrect, the feedback is a generic signal such as \texttt{TryAgain}. When the agent produces a correct answer, the episode terminates immediately. Consequently, no explicit positive confirmation (e.g., \texttt{Correct}) is ever added to the state history. The agent thus only receives unary feedback and must learn to revise its answers based solely on a history of failed attempts.

In practice, the prompt is constructed as a natural-language sequence concatenating all previous attempts and their feedback. For example:
\begin{quote}
\small
\texttt{Question: What is the value of ...? \\
Attempt 1: [wrong answer] \\
Feedback: Try Again. \\
...\\
Attempt K: [correct answer]}
\end{quote}

This UFO mechanism enables us to transform static single-turn datasets into multi-turn interaction episodes without requiring structural changes, expert annotations, or execution environments. Thus, UFO allows multi-turn RL on LLMs with minimal supervision.

\subsection{Reinforcement Learning with Unary Feedback}

Given the MDP formulation and the UFO-based observation design, we optimize the agent using RL to learn revision-aware and multi-turn policies. Since the dataset only contains the final-answer accuracy and lacks ground-truth reasoning traces, supervised finetuning is not applicable. Reinforcement learning, in contrast, enables exploration of diverse reasoning strategies under sparse and delayed supervision.

We adopt Proximal Policy Optimization (PPO) to train the policy $\pi_\theta$, following prior work~\citep{hu2025openreasonerzeroopensourceapproach, ragen} which shows that a learned critic enables fine-grained value estimates and stabilizes optimization.
At each episode, the agent interacts with a problem over multiple rounds. At each turn $t$, it observes input $x_t$, generates an answer $a_t$, and receives a binary reward $r_t \in \{0, 1\}$. The resulting trajectory is defined as:
\begin{align}
\tau = \{(x_1, a_1, r_1), (x_2, a_2, r_2), \dots, (x_T, a_T, r_T)\},
\end{align}
where $T \leq T_{\text{max}}$ is the number of turns before success or termination. The objective is to maximize the expected return:
\begin{align}
\mathcal{J}^{\text{RL}}(\theta) = \mathbb{E}_{\tau \sim \pi_\theta} \left[ \sum_{t=1}^{T} r_t \right].
\end{align}
We apply PPO with a clipped surrogate objective. For each training batch, we estimate the advantage $\hat{A}_t$ using a baseline value function and update the policy as:
\begin{align}
\mathcal{L}^{\text{PPO}}(\theta) = \mathbb{E}_t \left[ \min \left( \frac{\pi_\theta(a_t \mid x_t)}{\pi_{\theta_\text{old}}(a_t \mid x_t)} \hat{A}_t,\; \text{clip}(\frac{\pi_\theta(a_t \mid x_t)}{\pi_{\theta_\text{old}}(a_t \mid x_t)} \hat{A}_t, 1 - \epsilon, 1 + \epsilon) \right) \right].
\end{align}
Crucially, the UFO design enables the policy to condition on the full history of failure signals, giving rise to context-sensitive behaviors such as error correction, elimination, and hypothesis refinement—capabilities that are difficult to elicit through static supervision alone.

\subsection{Reward Design for Adaptive Reasoning}
\label{sec:reward_design}
Binary correctness signals offer a minimal form of supervision, but they could induce suboptimal behavior such as blind trial-and-error or repeated guesses. To encourage more efficient and reflective reasoning, we introduce a trajectory-level \textit{reward decay} with \textit{repetition penalty}. Reward decay encourages \textbf{minimality} by favoring trajectories that reach correct answers in fewer turns, thereby promoting concise and purposeful reasoning, while the repetition penalty promotes \textbf{diversity} by penalizing repetitive answers and encouraging the model to explore alternative reasoning strategies upon failure.

Formally, reward decay promotes early success by assigning exponentially diminishing rewards to correct answers produced at later turns:
\begin{align}
R_t = 
\begin{cases}
\gamma^t, & \text{if } a_t \text{ is correct}, \\
0, & \text{otherwise},
\end{cases}
\end{align}
where $\gamma \in (0, 1)$ is a decay factor that favors solving the problem in fewer turns.

We define the repetition penalty based on the number of \textit{effective answers}. Let $T$ denote the number of turns in the episode, and $E(\tau)$ be the number of effective answers in the trajectory $\tau$. We define a normalized penalty term:
\begin{align}
\text{Penalty}(\tau) = \lambda \cdot \left(1 - \frac{E(\tau)}{T} \right),
\label{eq:penalty}
\end{align}
where $\lambda > 0$ is a tunable penalty weight, and $E(\tau)/T$ measures answer diversity. The penalty is maximized when all answers are identical.

Combining above components, the trajectory-level reward for RL training is defined as:
\begin{align}
R =
R_t - \text{Penalty}(\tau).
\end{align}
To improve stability, we apply a small penalty $\eta < 0$ for each malformed or missing output across turns.
To summarize, the reward is determined by the reasoning correctness at the last turn, answer diversity and format correctness of model answers across turns.

%% file: sections/03_experiments.tex
\section{Experiments}
\label{sec:exp}

\subsection{Setup}
\label{sec:exp_setup}
\paragraph{Dataset}
We conduct major experiments on the {MATH} subset of {MetaMathQA}~\citep{metamathqa} dataset (MMQ-Math), where data are augmented from the MATH~\citep{hendrycks2021measuringmathematicalproblem} dataset. This environment provides math questions with adequate difficulty, enabling us to observe and analyze its reasoning emergence. We also select eight other widely-used datasets in four different domains to evaluate the generalization ability of UFO: TheoremQA~\citep{theoremqa} evaluates formal mathematics understanding through questions about theorem statements and proofs. GSM8K\citep{gsm8k} focuses on grade-school level arithmetic reasoning. GPQA\citep{gpqa} tests graduate-level physics understanding. MMLU-STEM\citep{MMLU} is a curated STEM-focused benchmark derived from fifteen scientific and technical subjects within MMLU. HotPotQA~\citep{hotpotqa} tests multi-hop factual reasoning across Wikipedia passages, while ConcurrentQA~\citep{concurrentqa} focuses on temporal and causal reasoning in concurrent event settings. MMLU~\citep{MMLU} assesses general-knowledge proficiency across fifty-seven academic subjects, and its extension, MMLU-Pro~\citep{MMLUPro} covers more specialized expert domains.

\paragraph{Training Settings}
We use Qwen-2.5-3B-Instruct~\citep{qwen25} with {PPO} for $200$ optimization steps on A100 GPUs as main training setting. Each batch samples $P{=}8$ prompts, with $N{=}16$ rollouts per prompt. During training, we experiment with three distinct maximum number of turns per episode, setting $T_{\text{max}}$ to 1, 5, and 10, respectively. For the validation phase, $T_{\text{max}}$ is fixed at 5 turns. In both training and validation, episodes are limited to a maximum of 10 actions in total. Policy updates use PPO with GAE parameters $(\gamma,\lambda)=(1.0,1.0)$, Adam with $\beta=(0.9,0.999)$, entropy coefficient $10^{-3}$. We apply the same training setup to four additional models: Qwen2.5-1.5B-Instruct, Qwen2.5-7B-Instruct, LLaMA3.2-1B-Instruct~\citep{llama3_2_blog}, and LLaMA3.2-3B-Instruct, to ensure consistent comparison across architectures and scales.
    
\paragraph{Baseline and Metrics}
We compare our method\textbf{ UFO} against a single-turn PPO-trained model using parallel sampling. For each problem, the baseline generates $k$ independent responses in parallel and is evaluated using standard \texttt{Pass@k} metric. In contrast, our multi-turn model generates responses sequentially with unary feedback after each attempt, and is evaluated using both \texttt{Succ@k} and \texttt{AvgTurns}. Success is recorded if any of the 5 responses is correct. We also conduct ablation studies with different maximum interaction turns ($T_{\text{max}}$) to further analyze the effect of multi-turn training.

We report the following complementary metrics to assess both effectiveness and efficiency.

\begin{itemize}[topsep=0pt, itemsep=2pt, parsep=0pt, leftmargin=10pt]
    \item \textbf{Pass@k (Single-turn baseline).}~
    The proportion of problems for which at least one of the k parallel completions is correct. This metric reflects performance when no feedback is used during generation.
    \item \textbf{Succ@k (Multi-turn model)}~~
This metric measures the percentage of problems solved within a fixed number of interaction turns. Let $\tau_j$ be the number of turns the agent takes to solve problem $q_j$, or $\infty$ if it fails. We have:
\begin{align}
\text{Succ@k} = \frac{1}{N} \sum_{j=1}^{N} \mathbbm{1}[\tau_j \le k].   
\end{align}
We report \texttt{Succ@1} for single-turn performance, and \texttt{Succ@5/10} to reflect multi-turn capability.
\item \textbf{Average Number of Turns (Multi-turn model)}~~
To evaluate interaction efficiency, we report the average number of turns the agent takes to solve each problem:
$\text{AvgTurns} = \tfrac{1}{N}\sum_{j=1}^{N} T_{j}$. $T_j$ denotes the number of interactive turns taken for problem $q_j$. This metric reflects how efficiently the agent reaches a solution, accounting for retries and step-wise refinement across multi-turn episodes.
\end{itemize}

%% file: sections/04_analysis.tex
\begin{figure}[t]
    \centering
    \begin{minipage}[t]{0.49\linewidth}
        \centering
        \includegraphics[width=\linewidth]{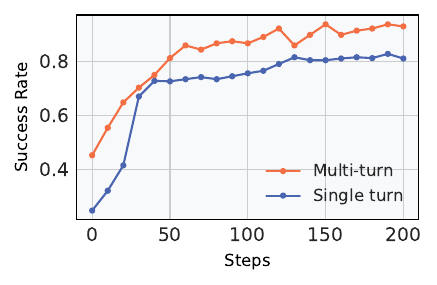}
        \caption{Multi-turn (5-turn) RL significantly outperforms single-turn baseline, achieving higher success rates (Pass @ 5) with similar inference cost.}
        \label{fig:compare_baseline}
    \end{minipage}
    \hfill
    \begin{minipage}[t]{0.49\linewidth}
        \centering
        \includegraphics[width=\linewidth]{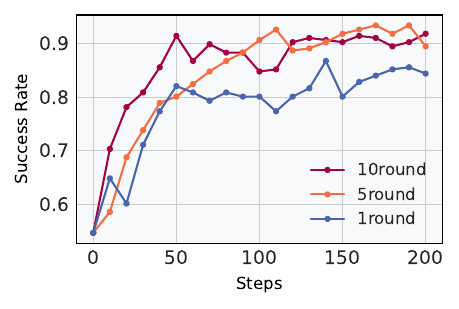}
        \caption{Performance comparison when evaluating with 5 turns after training with different maximum turns (1, 5, and 10). Training with 5 turns yields the best performance, while increasing to 10 turns offers no significant gain.}
        \label{fig:1051}
        \end{minipage}
\end{figure}

\begin{figure}[t]
    \centering
    \includegraphics[width=0.98\linewidth]{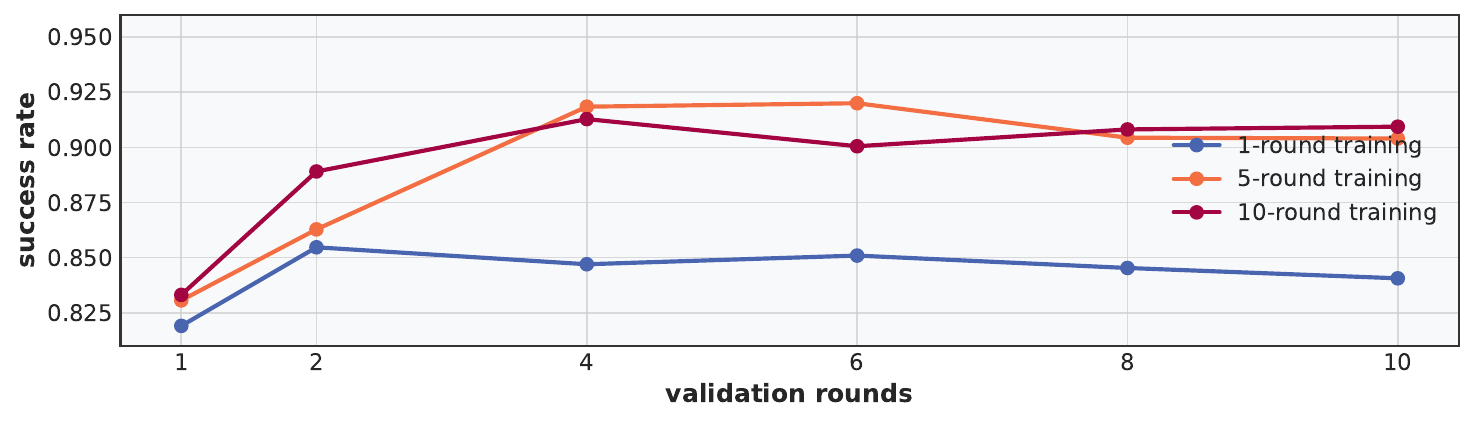}
    \caption{\textbf{Validation performance (\texttt{Succ@k}) of models trained with different roll-out turns under varying inference-time turn budgets.} Multi-turn training (5 or 10 turns) consistently yields higher success rates across all inference turn budgets, including $k=1$, indicating better generalization even to single-turn reasoning. }
    \label{fig:validation_rounds}
\end{figure}

\begin{table*}[t]
\centering
\caption{5-turn success rate across different tasks and training settings.}
\vspace{10pt}
\scriptsize 
\setlength{\tabcolsep}{2pt}
\begin{tabular}{lccccccccc}
\toprule
& \multicolumn{3}{c}{Math} & \multicolumn{2}{c}{STEM} & \multicolumn{2}{c}{QA} & \multicolumn{2}{c}{General} \\
\cmidrule(lr){2-4} \cmidrule(lr){5-6} \cmidrule(lr){7-8} \cmidrule(lr){9-10}
Model & MMQ-Math & TheoremQA & GSM8k & GPQA & MMLU-STEM & HotpotQA & ConcurrentQA & MMLU & MMLU-Pro \\
\midrule
\multicolumn{10}{l}{\textbf{Qwen2.5-1.5B-Instruct}} \\
Base Model w/o RL            & 10.9 & 11.7 & 26.6 & 21.9 & 62.5 & 2.4 & 3.1 & 52.3 & 35.2 \\
RL on MMQ-Math               & 74.8 & 20.1 & 84.7 & 22.7 & \textbf{65.5} & 19.2 & \textbf{9.5} & 43.8 & \textbf{34.8} \\
\rowcolor{gray!15}
+5turn UFO                   & \textbf{83.6} & \textbf{26.8} & \textbf{88.1} & \textbf{27.3} & 64.8 & \textbf{22.6} & \textbf{9.5} & \textbf{60.9} & \textbf{34.8} \\
\midrule
\multicolumn{10}{l}{\textbf{Qwen2.5-3B-Instruct}} \\
Base Model w/o RL            & 52.3 & 28.3 & 68.0 & 51.6 & 75.8 & 7.8 & 3.9 & 75.2 & 42.2 \\
RL on MMQ-Math               & 79.7 & 32.0 & 93.0 & 50.1 & 77.6 & 19.5 & 12.9 & 66.8 & 48.3 \\
\rowcolor{gray!15}
+5turn UFO                   & \textbf{88.5} & \textbf{40.8} & \textbf{95.3} & 52.3 & 87.5 & 26.6 & 15.2 & \textbf{85.2} & \textbf{60.9} \\
RL on HotQA                  & 72.4 & 31.8 & 89.1 & 48.4 & 81.3 & 38.3 & \textbf{16.8} & 71.5 & 49.3 \\
\rowcolor{gray!15}
+5turn UFO                   & 72.7 & 29.2 & 85.0 & \textbf{57.8} & \textbf{88.3} & \textbf{44.2} & \textbf{16.8} & 76.6 & 48.9 \\
\midrule
\multicolumn{10}{l}{\textbf{Qwen2.5-7B-Instruct}} \\
Base Model w/o RL            & 56.4 & 32.1 & 56.3 & \textbf{62.5} & 83.6 & 13.3 & 4.7 & 72.3 & 64.1 \\
RL on MMQ-Math               & 85.1 & 33.6 & 95.2 & 50.8 & \textbf{84.8} & 26.3 & 14.1 & 73.4 & 52.3 \\
\rowcolor{gray!15}
+5turn UFO                   & \textbf{93.0} & \textbf{42.1} & \textbf{96.8} & 56.9 & \textbf{84.8} & \textbf{28.6} & \textbf{16.4} & \textbf{80.5} & \textbf{58.8} \\
\midrule
\multicolumn{10}{l}{\textbf{Llama3.2-1B-Instruct}} \\
Base Model w/o RL*           & 2.3 & 2.3 & 1.6 & 1.6 & 4.6 & 0.8 & 0.8 & 3.9 & 2.3 \\
RL on MMQ-Math               & 53.9 & 21.1 & 52.3 & 20.3 & 57.0 & 19.5 & 0.8 & 57.8 & \textbf{32.8} \\
\rowcolor{gray!15}
+5turn UFO                   & \textbf{64.8} & \textbf{26.8} & \textbf{56.3} & \textbf{26.6} & \textbf{60.2} & \textbf{21.1} & \textbf{1.6} & \textbf{66.4} & \textbf{32.8} \\
\midrule
\multicolumn{10}{l}{\textbf{Llama3.2-3B-Instruct}} \\
Base Model w/o RL            & 50.8 & 20.3 & 48.4 & 47.7 & 77.3 & 29.7 & 6.0 & 65.6 & 49.2 \\
RL on MMQ-Math               & 86.7 & 24.2 & 92.2 & 46.9 & 78.1 & \textbf{44.5} & 13.3 & 71.1 & 60.9 \\
\rowcolor{gray!15}
+5turn UFO                   & \textbf{92.2} & \textbf{32.0} & \textbf{93.8} & \textbf{50.8} & \textbf{82.0} & 39.8 & \textbf{14.8} & \textbf{82.8} & \textbf{66.4} \\
\bottomrule
\end{tabular}
\label{tab:5turn-success}
\end{table*}

\begin{figure}[t]
    \centering
    \includegraphics[width=0.98\linewidth]{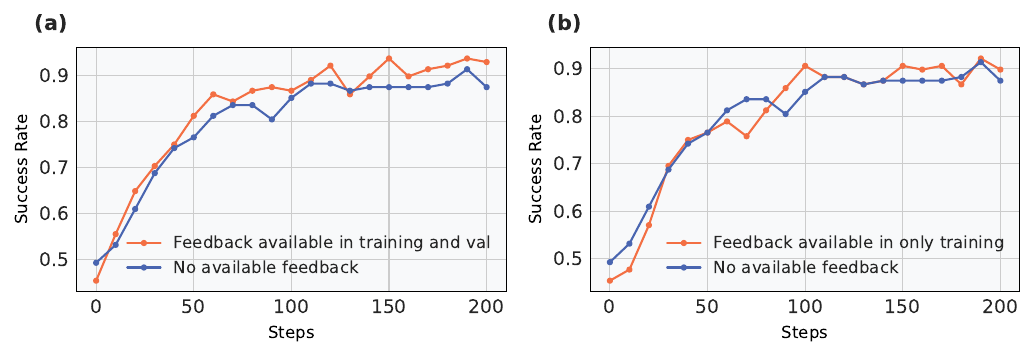}
    \caption{\textbf{Comparison of success rate with multi-turn setting.} (a) with feedback prompt in both training and validation compared to blank prompt; (b) with feedback prompt only in training compared to blank prompt.}
    \label{fig:compare}
\end{figure}

\subsection{Experimental Results and Findings}
\linespread{0.97}\selectfont
In this section, we present empirical findings that address three central questions in our study of multi-turn reinforcement learning with unary feedback:

\begin{enumerate}[noitemsep, topsep=0pt, leftmargin=*]
    \item Section~\ref{sec:4.2.1}: Does multi-turn RL unlock stronger reasoning than single-turn training?
    \item Section~\ref{sec:4.2.2}: Can models effectively revise their answers from sparse feedback alone?
    \item Section~\ref{sec:4.2.3}: How do reward shaping strategies impact reasoning efficiency and diversity?
\end{enumerate}

We explore each question in the following subsections, with quantitative analyses and ablation studies. Additional qualitative examples and robustness checks are included in the Appendix.

\subsubsection{Multi-turn RL Unlocks Higher Upper Bound of LLM Reasoning}

\label{sec:4.2.1} 
We compare models trained with multi-turn RL against single-turn PPO baselines, using \texttt{Succ@5} on a held-out validation set evaluated at 21 checkpoints across 200 training steps. During validation, each agent is allowed up to 5 interaction turns per problem ($k=5$). 

As shown in Figure~\ref{fig:compare_baseline}, multi-turn training consistently outperforms the single-turn baseline, \textbf{achieving up to 14\% higher success rate} with comparable inference cost. This highlights the benefit of iterative revision under sparse feedback.

Furthermore, we conduct additional experiments comparing various multi-turn training budgets ($T_{\text{max}} = 1, 5, 10$) while consistently using a 5-turn validation setup. Findings presented in Figure~\ref{fig:1051} demonstrate that larger training budgets yield enhanced performance relative to the single-turn baseline. Notably, both the $T_{\text{max}}=10$ and $T_{\text{max}}=5$ configurations deliver \textbf{more than a 6\% relative improvement} over single-turn training at their peak, clearly emphasizing the benefits of multi-turn training.

To validate the robustness of these improvements, we expand our analysis by evaluating peak-performing models trained with $T_{\text{max}} \in \{1, 5, 10\}$ across varied inference-time interaction budgets ($k \in \{1, 2, 4, 6, 8, 10\}$). The results illustrated in Figure~\ref{fig:validation_rounds} reinforce previous observations, consistently showing superior Succ@$k$ performance by models trained under multi-turn conditions. Intriguingly, these improvements are observable even at the lowest inference budget ($k=1$), suggesting that multi-turn training enhances not only iterative performance but also generalizes well to single-shot scenarios. 

\begin{figure}[t]
    \centering
    \includegraphics[width=0.9\linewidth]{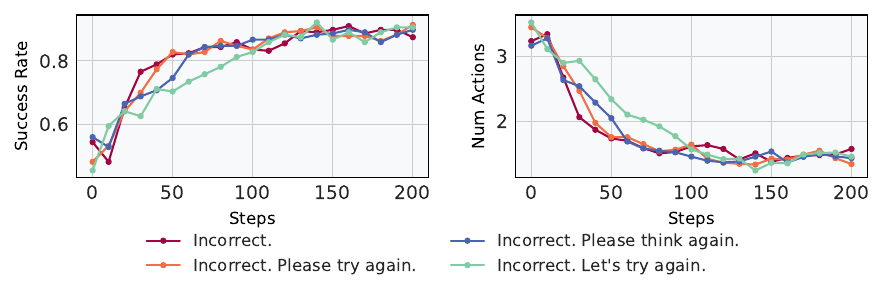}
    \caption{\textbf{Validation under different verbal feedback prompts.} Success rates and action counts remain consistent across all variants, demonstrating UFO's robustness to various prompts.}
    \label{fig:prompt}
\end{figure}

\begin{figure}[t]
    \centering
    \includegraphics[width=0.9\linewidth]{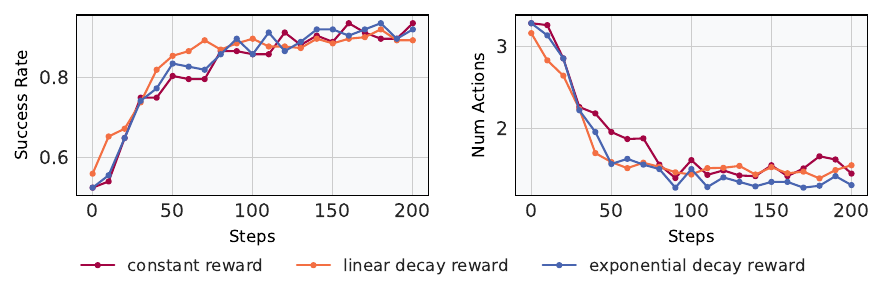}
    \caption{\textbf{Comparison of reward shaping strategies.} While constant, linear decay, and exponential decay schedules achieve similar success rates (left), exponential decay consistently leads to fewer actions per episode (right), indicating more efficient problem solving with less external supervision.}
    \label{fig:reward_design}
\end{figure}

Table~\ref{tab:5turn-success} summarizes three key findings across five models and nine datasets. First, applying 5-turn UFO consistently improves performance over single-turn RL on the same task, confirming its in-domain effectiveness. Second, UFO-trained models generalize reliably to new domains, with consistent gains observed across four broad categories: math, STEM, QA, and general knowledge. Third, math-trained models tend to generalize more strongly than QA-trained ones, possibly due to the more structured and logically consistent nature of mathematical reasoning. 
These trends hold across two families of open-source models, Qwen2.5 and Llama3.2, spanning from 1B to 7B in scale, suggesting the effectiveness of UFO are robust to both model architecture and size. \textit{We note that LLaMA3-1B shows relatively lower base performance compared to official numbers, possibly due to format-following issues.}

These trends suggest that by enabling models to explore alternative reasoning paths and revise prior failures based on sparse feedback, UFO provides a principled path toward robust multi-turn and cross-task generalization.

\subsubsection{Multi-turn Setting Enables LRMs to Revise From Feedback}
\label{sec:4.2.2}

The multi-turn setting enables agents to engage repeatedly with each prompt (up to $T_{\text{max}}$ turns), thereby constructing richer and more informative interaction trajectories from the same training data. This enhanced utilization of feedback is hypothesized to extract more meaningful learning signals per problem, potentially improving solution quality and accelerating convergence, especially in data-limited contexts.

To empirically validate that LRMs can be improved effectively utilizing conversational feedback for revision, we compare 5-turn training scenarios with and without explicit feedback prompts. Results presented in Figure~\ref{fig:compare}(a) support this hypothesis, \textbf{demonstrating an over 8\% peak performance improvement when explicit feedback is provided.}

An additional analysis with feedback prompt only in training (Figure~\ref{fig:compare}(b)) reveals performance improvement as well. This suggests that multi-turn training can even intrinsically enhance model reasoning capabilities.

Finally, our robustness analysis in the Figure~\ref{fig:prompt} shows that the effectiveness of this approach is preserved across a range of prompt formulations, underscoring its practical applicability in real-world scenarios.

\subsubsection{Reward Shaping Encourages Efficient Problem Solving}
\label{sec:analysis_reward_decay}
\label{sec:4.2.3}

We investigate how different reward schedules influence the agent's learning behavior, particularly in encouraging early success versus allowing extended exploration. All schedules define a reward $r(n)$ based on the turn index $n$ when the first correct answer is produced, with $n \in \{1, \dots, T_{\text{max}}\}$.

We define and evaluate three distinct reward schedules. Following the formulas proposed in Section~\ref{sec:reward_design}, we compare three approaches: (1) \textbf{Exponential Decay:} $r_{\text{exp}}(n) = \gamma^n$ (with $\gamma=0.5$), (2) \textbf{Linear Decay:} $r_{\text{lin}}(n) = \max(0, 1 - 0.2(n-1))$, (3) \textbf{Constant Reward:} $r_{\text{const}}(n) = 1.$
All schedules operate for $n \in {1, \dots, T_{\text{max}}}$.
The agent's objective remains to maximize the expected cumulative reward.

Experimental validation (Figure~\ref{fig:reward_design}) confirms that \textbf{exponential reward decay notably reduces the mean number of actions by roughly 10\%}, without sacrificing overall success rates. This reduction in action count suggests that the exponential decay schedule encourages the model to engage in more profound self-reflection and systematic thinking before generating a response.

\begin{figure}[!th]
    \centering
    \includegraphics[width=0.98\linewidth]{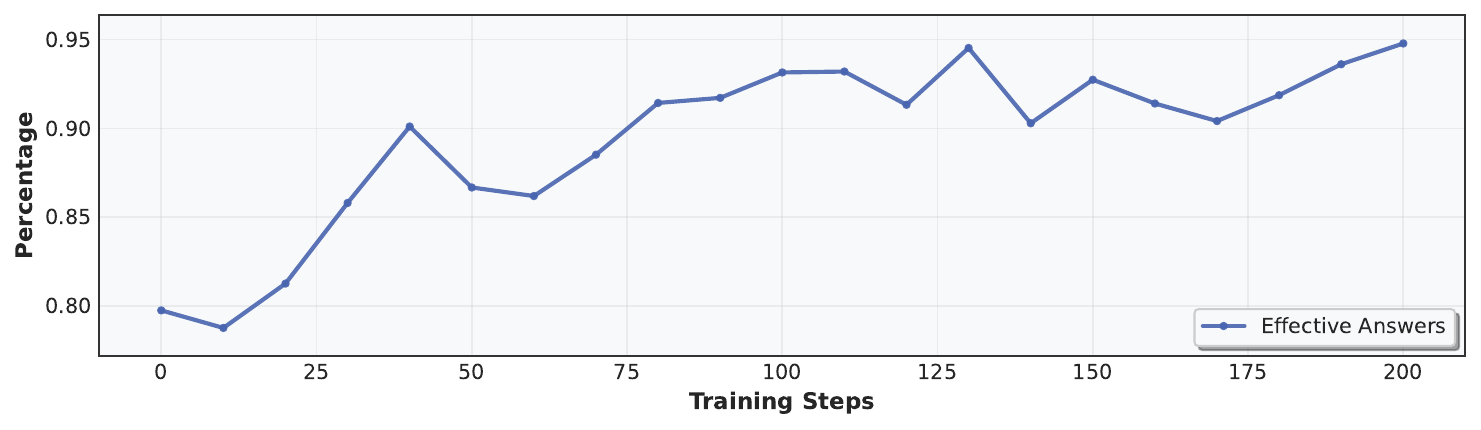}
    \caption{\textbf{Proportion of effective answers over training.} The upward trend suggests improved diversity across turns, which reduces penalty from repeated responses and contributes to higher overall rewards.}
    \label{fig:non-repetitive}
\end{figure}

Considering the normalized penalty term in our experiment (Equation~\ref{eq:penalty}),
we count the number of non-repetitive answer for each validation round, as shown in Figure~\ref{fig:non-repetitive}. The percentage \textbf{increases from 80\% to over 90\%}, suggesting that the model performs better in the later stages of training as the model learned to explore and self-reflect.

%% file: sections/05_related_work.tex
\section{Related Work}
\paragraph{Enhancing LLM Reasoning with Test‑Time Search and Parameter‑Efficient Training.} \textbf{Test-time reasoning frameworks} keep model weights frozen yet boost performance: 
Graph-of-Thought~\citep{peng2023got}, Reflexion~\citep{shinn2023reflexion}, 
Monte Carlo Tree Self-Refine~\citep{zhang2024mcts}, Self-Refine~\citep{madaan2023selfrefine}, 
CRITIC~\citep{gou2023critic} and memory-augmented agents such as 
POEM~\citep{do2024poem} and Larimar~\citep{liu2024larimar} rely on search, self-feedback or episodic memory \emph{without} updating model parameters. 

\paragraph{Training-time optimisation methods}, in contrast, adjust the policy itself.  
RLHF~\citep{christiano2017deep, instructgpt} and its low-cost variant RLAIF~\citep{lee2023rlaif} align models to preference data; scalable-oversight \emph{debate} protocols explore alignment with weak judges~\citep{kenton2024scalable}.  
Lightweight objectives such as Direct Preference Optimisation (DPO)~\citep{rafailov2023dpo}, Parameter-Efficient RLHF (PERLHF)~\citep{sidahmed2024perlhf} and Self-Play Fine-Tuning (SPIN)~\citep{chen2024spin} further cut roll-out cost, while hierarchical ArCHer~\citep{archer} tackles long-horizon credit assignment.  
Benchmarks like UNO Arena~\citep{qin2024unoarena} expose the strengths and weaknesses of both families in stateful, multi-turn settings.

\paragraph{Multiturn training for LLMs.}
Multiturn training for large language models (LLMs) has been explored across benchmarks, optimization methods, and architectural innovations. Evaluation benchmarks such as LMRL-Gym~\citep{abdulhai2023lmrlgym} and MT-Eval~\citep{kwan2024mteval} assess LLMs’ abilities to maintain consistency, follow instructions, and exhibit coherent planning across dialogue turns. On the training side, several works extend RLHF to multiturn scenarios by optimizing rewards over full dialog trajectories, including regression-based value estimation~\citep{gao2025refuel}, hierarchical actor-critic methods~\citep{archer}, and direct preference modeling with trajectory normalization~\citep{shi2024dmpo}. Further improvements integrate execution feedback in mathematical agents~\citep{xiong2024mathagent}, while early efforts focus on optimizing full-dialogue preferences~\citep{shani2024multiturn}. Additional frameworks explicitly model long-horizon collaboration through multiturn-aware rewards, as seen in CollabLLM~\citep{collabllm2025} and the modular self-evolving architecture of RAGEN~\citep{ragen}. Beyond RL-based approaches, parameter-efficient fine-tuning methods such as Baize~\citep{xu2023baize} demonstrate strong multiturn capabilities via LoRA adaptation on self-chat data. A recent survey~\citep{zhang2025survey} provides a comprehensive taxonomy covering these approaches, including optimization techniques, memory strategies, and evaluation protocols.

%% file: sections/06_imp_ccl.tex
\vspace{-5pt}
\section{Conclusions and Limitations}

\label{sec:conclusions}

In this work, we highlight a critical limitation of current single-turn RL training: its tendency to impair multi-turn reasoning by promoting repetitive and shallow responses. To address this, we propose \textit{Unary Feedback as Observation (UFO)}, a simple yet effective method that integrates minimal feedback into existing RL pipelines. By explicitly incorporating prior interaction history, UFO enables models to engage in exploration and self-reflection across multiple attempts, leading to deeper reasoning and improved adaptability. Our experiments show a 14\% gain in multi-turn accuracy while preserving single-turn quality. Additionally, we demonstrate that incorporating reward decay and repetitive penalty encourages diverse reasoning, self-correction and more thoughtful response patterns. Our approach is lightweight, generalizable, and easily applicable to existing datasets. A limitation of our work is its primary focus on relatively small models, leaving its generalizability to larger scales for future investigation. 

\section{Acknowledgements}
We thank the DeepSeek team for providing the DeepSeek-R1 model and early conceptual inspirations. We are grateful to the veRL team for their infrastructure support and the RAGEN team for their multi-turn RL framework. We thank Anna Zhao for their valuable contribution in helping revising manuscripts.

%% file: sections/07_appendix_supplementary.tex
\appendix
\section{Extended Background of Reinforcement Learning in LLMs}
\label{app:background_rl}

Reinforcement Learning (RL) enables large language models to improve through interaction and reward feedback. The general RL objective maximizes the expected reward over sampled responses:
\begin{equation}
    J(\theta) = \mathbb{E}_{x \sim \mathcal{D},\, y \sim \pi_\theta(\cdot \mid x)} \left[ R(x, y) \right],
\end{equation}
where \( \pi_\theta \) is the model policy, \( x \) is the input prompt, \( y \) is the generated output, and \( R(x, y) \) is a scalar reward assessing response quality.

A widely adopted method for RL fine-tuning is Proximal Policy Optimization (PPO)~\citep{ppo}, which stabilizes training by clipping the likelihood ratio between the new and old policies. The ratio is defined as:
\begin{equation}
    \rho_t(\theta) = \frac{\pi_\theta(y_t \mid x_t)}{\pi_{\theta_{\text{old}}}(y_t \mid x_t)}.
\end{equation}
The PPO objective minimizes over the clipped surrogate advantage:
\begin{equation}
    J_{\text{PPO}}(\theta) = \mathbb{E}_t \left[ \min\left( \rho_t A_t, \hat{\rho}_t A_t \right) - \beta D_{\mathrm{KL}} \right],
\end{equation}
where \( \hat{\rho}_t = \text{clip}(\rho_t, 1 - \epsilon, 1 + \epsilon) \), and \( A_t \) is the advantage function estimating how much better \( y_t \) is than the baseline under prompt \( x_t \).

For advantage estimation, Generalized Advantage Estimation (GAE)~\citep{schulman2015GAE} is often used:
\begin{equation}
    A_t^{\mathrm{GAE}(\gamma, \lambda)} = \sum_{l=0}^{\infty} (\gamma \lambda)^l \delta_{t+l}, \quad \text{with} \quad \delta_t = r_t + \gamma V(x_{t+1}) - V(x_t),
\end{equation}
where \( (\gamma, \lambda) \) trade off bias and variance.

More recently, DeepSeekMath \citep{deepseekmath} and DeepSeek-R1 \citep{deepseekr1} adopts Group Relative Policy Optimization (GRPO), a RL method that samples a set of outputs \( \{y_i\}_{i=1}^G \) for each prompt \( x \), and optimizes:
\begin{equation}
    J_{\text{GRPO}}(\theta) = \mathbb{E}_{x, \{y_i\}} \left[ J_{\text{group}}(\theta) \right],
\end{equation}
with
\begin{equation}
    J_{\text{group}}(\theta) = \frac{1}{G} \sum_{i=1}^G \min\left( \rho_i A_i, \hat{\rho}_i A_i \right) - \beta D_{\mathrm{KL}},
\end{equation}
where the advantage \( A_i \) is computed using a reward-normalized baseline:
\begin{equation}
    A_i = \frac{r_i - \text{mean}(\{r_j\})}{\text{std}(\{r_j\})}.
\end{equation}
This avoids dependency on value networks and uses rule-based or environment-specific rewards \( r_i \), making it well-suited for reasoning tasks where explicit heuristics can guide learning. GRPO has shown to induce emergent multi-step reasoning behavior across domains.

\section{Details on the Theoretical Analysis on Repetition Pattern}
\label{sec:app-repetition-theory}

We provide a detailed proof on how peaked, low-entropy output distribution from RL training \citep{cui2025entropymechanismreinforcementlearning, yue2025doesreinforcementlearningreally} can lead to high repetition in model multi-turn behavior.

\paragraph{Preliminaries.} Let $q(y \mid x)$ denote the model's output distribution given input $x$. We introduce the following definitions:

\begin{definition}[Collision Probability]
The collision probability of $q(y \mid x)$ is defined as:
\begin{equation}
\mathrm{Coll}(q) := \sum_y q(y \mid x)^2.
\end{equation}
This is the probability that two i.i.d.\ samples from $q$ yield the same answer: $\Pr[A_i = A_j] = \mathrm{Coll}(q)$.
\end{definition}

\begin{definition}[Entropy]
The Shannon entropy of $q(y \mid x)$ is:
\[
\mathcal{H}(q) := -\sum_y q(y \mid x) \log q(y \mid x).
\]
Lower entropy corresponds to a more peaked distribution.
\end{definition}

\begin{definition}[Expected Number of Duplicate Pairs]
Given $k$ i.i.d.\ samples $\{A_1, \dots, A_k\} \sim q(\cdot \mid x)$, the expected number of duplicate pairs is:
\[
\mathbb{E}[\mathrm{DupPairs}] = \binom{k}{2} \cdot \mathrm{Coll}(q).
\]
\end{definition}

\paragraph{Repetition Under Sequential Sampling.}
\begin{proposition}
\label{prop:collision-entropy}
Let $A_1, \dots, A_k \sim q(\cdot \mid x)$ be sampled sequentially.
Since single-turn RL does not guarantee any multi-turn capability, we simplify the assumption that the policy is static and does not update based on prior turns, i.e., the answer at each turn $t$ is sampled from the same fixed distribution $q(\cdot | x)$. The probability of generating a duplicate answer is the collision probability, which is lower-bounded by:
\[
\Pr[A_i = A_j] \ge \exp(-\mathcal{H}(q)),
\]
where $\mathcal{H}(q)$ denotes the Shannon entropy of the base distribution $q(\cdot \mid x)$.
\end{proposition}

\begin{proof}
Let \( Y \sim q(\cdot \mid x) \), and define the collision probability as:
\[
\mathrm{Coll}(q) = \mathbb{P}[A_i = A_j] = \sum_y q(y)^2 = \mathbb{E}_{Y \sim q}[q(Y)].
\]
By Jensen’s inequality applied to the concave function \( \log \), we have:
\[
\log \mathbb{E}_{Y}[q(Y)] \ge \mathbb{E}_{Y}[\log q(Y)] = -\mathcal{H}(q),
\]
which implies:
\[
\mathrm{Coll}(q) \ge \exp(-\mathcal{H}(q)).
\]
\end{proof}

\begin{remark}[On the Tightness of the Bound]
The lower bound $\mathrm{Coll}(q) \ge \exp(-\mathcal{H}(q))$ is tight in the following cases:
\begin{itemize}
    \item When $q$ is uniform on a support of size $n$, i.e., $q(y) = 1/n$, then $\mathcal{H}(q) = \log n$ and $\mathrm{Coll}(q) = 1/n$, achieving equality.
    \item When $q$ is a delta distribution (i.e., concentrated on one point), then $\mathcal{H}(q) = 0$ and $\mathrm{Coll}(q) = 1$.
\end{itemize}
We assume $\log q(y)$ is only computed where $q(y) > 0$, so the result still holds for distributions with zero-probability points.

A tighter bound can be obtained using the Rényi–Shannon inequality:
\[
\mathrm{Coll}(q) = e^{ -\mathcal{H}(q) - D_{\mathrm{KL}}(q \parallel u)},
\]
where $u$ is the uniform distribution on the support of $q$. Our stated bound omits the KL divergence for simplicity and interpretability.
\end{remark}

\noindent
This demonstrates that a low-entropy model is mathematically guaranteed to have a higher floor for its repetition rate. Thus, for a static agent that does not learn from feedback, repetition is not an accidental bug but a predictable outcome of the low-entropy distributions created by standard RL.

\section{Theoretical Analysis on Advantages of Sequential Policies}
\label{app:theory-ufo}
We model multi-turn reasoning as a finite-horizon Markov Decision Process (MDP), where the state at step \( t \) is given by
\[
s_t = (q, a_1, f_1, \dots, a_{t-1}, f_{t-1}),
\]
with input question \( q \), previous answers \( a_i \), and corresponding feedback \( f_i \). 

We represent policies as functions \( \pi_t(a \mid s_t) \) that produce an action \( a \) based on the interaction history \( s_t \). In our setting, the \textit{single-turn RL} paradigm corresponds to a \textbf{parallel policy}, where the model samples multiple answers independently from a fixed distribution \( \pi(a \mid q) \) without conditioning on feedback. In contrast, our \textit{UFO} (Unary Feedback as Observation) operates as a \textbf{sequential policy}, where the action at each step is conditioned on the full history and thus can adapt dynamically.

We formalize the distinction as follows:
\begin{itemize}
    \item \textbf{Parallel policies} \( \pi \in \Pi_{\text{par}} \): sample answers i.i.d.\ from a fixed policy \( \pi(a \mid q) \), without using feedback.
    \item \textbf{Sequential policies} \( \pi \in \Pi_{\text{seq}} \): choose actions based on the full state \( s_t \), enabling feedback-driven refinement.
\end{itemize}

Every parallel policy is a special case of a sequential policy that ignores interaction history. Formally, for any \( \pi(a \mid q) \), we can construct \( \pi_t(a \mid s_t) = \pi(a \mid q) \) for all \( t \), implying
\[
\Pi_{\text{par}} \subseteq \Pi_{\text{seq}}.
\]
Since \( \Pi_{\text{par}} \subseteq \Pi_{\text{seq}} \), we immediately have:
\[
\max_{\pi \in \Pi_{\text{seq}}} \mathbb{E}[\mathrm{Succ}@k] \ge \max_{\pi \in \Pi_{\text{par}}} \mathbb{E}[\mathrm{Succ}@k],
\]
where \(\mathrm{Succ}@k\) denotes the probability of producing a correct answer within \( k \) attempts.

Let \( p \) denote the success probability under a parallel policy. Then:
\[
\mathbb{P}_{\text{par}}[\text{success in } k \text{ turns}] = 1 - (1 - p)^k.
\]
For a sequential policy, let \( p'_t \) be the conditional success probability at step \( t \), which may depend on the state \( s_t \). The success probability is:
\[
\mathbb{P}_{\text{seq}}[\text{success in } k \text{ turns}] = 1 - \prod_{t=1}^k (1 - p'_t), \quad \text{where } p'_t \ge p.
\]
Sequential policies can eliminate previously failed answers by maintaining a rejection set \( \mathcal{H}_t \subset \mathcal{A} \) and enforcing:
\[
\pi_t(a \mid s_t) = 0 \quad \text{for all } a \in \mathcal{H}_t.
\]
This behavior approximates sampling without replacement and yields increasing conditional success rates:
\[
p'_t = \frac{p}{1 - \sum_{i=1}^{t-1} p_i} > p.
\]
Therefore, sequential policies such as UFO improve \(\mathrm{Succ}@k\) by adaptively avoiding prior failure modes. This formalizes the theoretical advantage of feedback-aware reasoning strategies over static single-turn RL.

\section{Detailed Evaluation under Multi-round Settings}

\begin{figure}[t]
    \centering
    \includegraphics[width=0.98\linewidth]{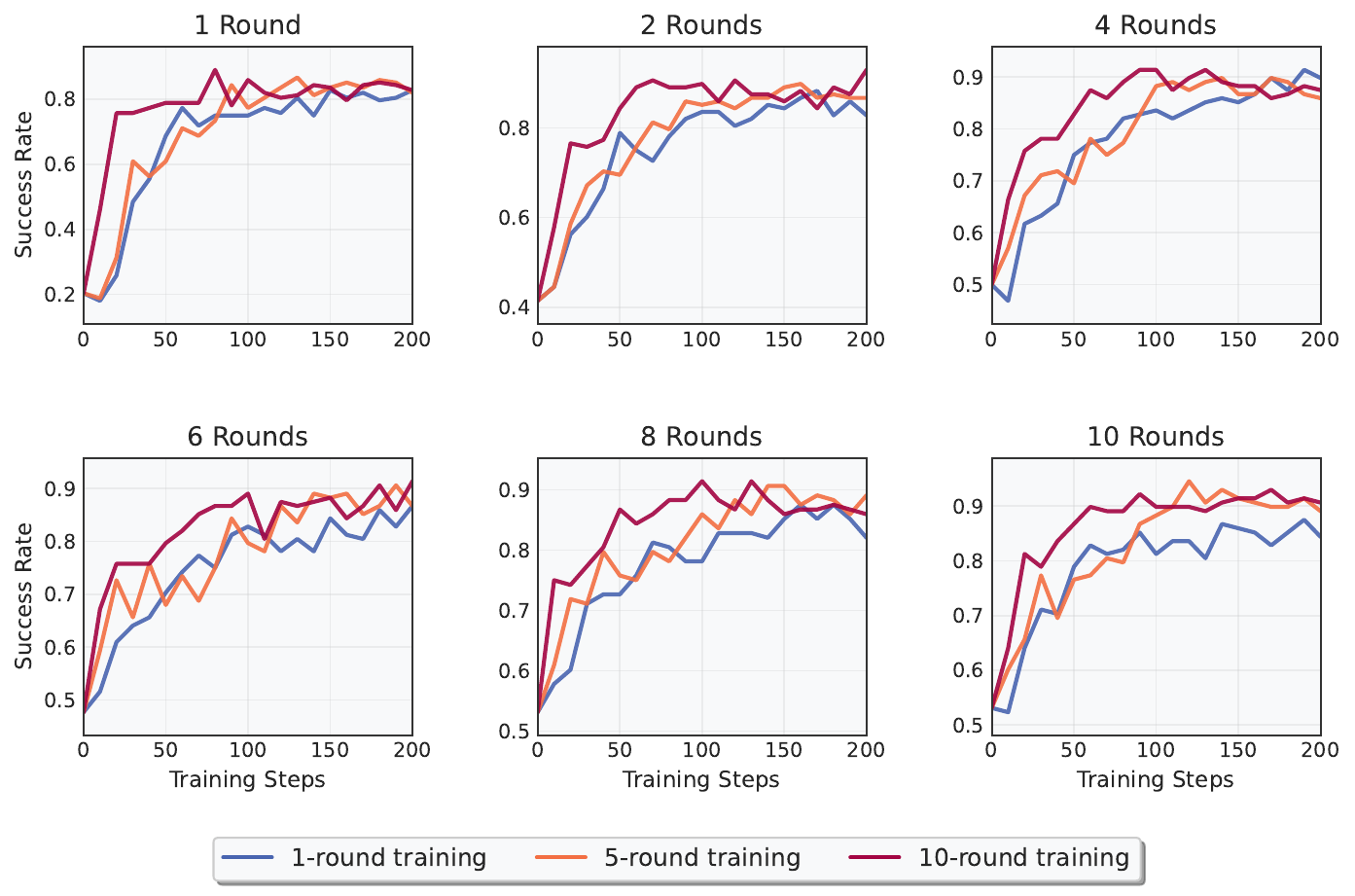}
    \caption{\textbf{Performance across different evaluation round settings.} Each subplot shows the success rate evaluated at $r$ rounds. While all methods perform similarly under 1-round evaluation, models trained with multi-round feedback (UFO) generalize significantly better to longer evaluation horizons.}
    \label{fig:enter-label}
\end{figure}

We illustrate a detailed analysis of how multi-round training improves generalization on long-horizon interactive reasoning. Figure~\ref{fig:enter-label} provides a comprehensive view of validation performance across all checkpoints, comparing models trained under 1-round, 5-round, and 10-round settings. Each curve represents evaluation success rates under a fixed number of evaluation rounds.

We observe that under 1-round evaluation (top-left), all training strategies achieve similar performance, suggesting that even single-turn training can suffice in this limited setting. However, as evaluation round count increases, the gap between single-round training and multi-round training becomes increasingly significant. In particular, models trained with 10-round UFO feedback consistently outperform the others under 6, 8, and 10-round evaluation, demonstrating more stable and generalizable behavior across turns.

These results support our core hypothesis: \textbf{unary feedback, when used as structured observation during training, enables better long-horizon generalization.} In contrast, models trained only with single-round interactions struggle to adapt to multi-turn dynamics, leading to degraded performance as the task horizon increases.

\section{Prompt Settings}
\subsection{Problem Solving Model Prompt}
We adopt a simple and structured prompt format for mathematical problem solving, following prior designs from~\citet{deepseekmath, qwen25}, with an extension to support multi-turn interactions. 
A key element of our prompt is the explicit \texttt{<think>} and \texttt{<answer>} separation, paired with an action budget (\texttt{Y}) and max length (\texttt{Z}). This guides the model to reason step-by-step while planning within a fixed turn horizon, improving controllability and alignment in multi-turn settings.
As shown in Box~\ref{box:prompt}, we present the prompt template used during training and evaluation.

\begin{tcolorbox}[breakable, colback=gray!2, colframe=black, boxrule=0.5pt, arc=2pt, width=\linewidth, title=Box 1: Model Prompt Template]
\refstepcounter{examplebox}\label{box:prompt}
\texttt{<|im\_start|>system\\
\{prompt\}\\
You're a helpful assistant.\\
<|im\_end|>\\
<|im\_start|>user\\
\{prompt\}\\
You are solving Math problems.\\
Turn X:\\
State:\\
(Question)\\
You have Y actions left. Always output: <think> [Your thoughts] </think> <answer> [your answer] </answer> with no extra text. Strictly follow this format. Max response length: Z words (tokens).\\
<|im\_end|>\\
<|im\_start|>assistant\\
... (This conversation pattern repeats for up to K turns) \\
<|im\_end|>
}\par
\end{tcolorbox}

\subsection{Feedback Provider Model Prompt}
We also present the prompt for the feedback provider that gives the problem-solving model more detailed feedback as follows.

\begin{tcolorbox}[breakable, colback=gray!2, colframe=black, boxrule=0.5pt, arc=2pt, width=\linewidth, title=Box 2: Tutor Prompt Template]
\refstepcounter{examplebox}\label{box:tutorprompt}
\texttt{<|im\_start|>system\\
You are a helpful math tutor.\\
<|im\_end|>\\
<|im\_start|>user\\
Problem: \{question\}\\
Student's answer: \{wrong\_answer\}\\
This answer is incorrect. Give a brief, encouraging hint (1--2 sentences) that helps the student reconsider their approach without revealing the correct answer. Focus on guiding them to check their work or think about the problem differently.\\
Response format: Just the hint, no extra formatting.\\
<|im\_end|>\\
<|im\_start|>assistant\\
...\\
<|im\_end|>}
\end{tcolorbox}

\subsection{TheoremQA Prompt Format}
The TheoremQA environment follows a similar prompting structure as the MMQ-Math, with an additional image token placed at the beginning of the question when a picture is present. This enables compatibility with multimodal pipelines, where the image is processed separately while the text prompt includes a placeholder token to signal its presence.

As shown in Box~\ref{box:theoremqa}, the image token \texttt{<image>} is placed on a new line above the question if an image is available.

\begin{tcolorbox}[breakable, colback=gray!2, colframe=black, boxrule=0.5pt, arc=2pt, width=\linewidth, title=Box 3: TheoremQA Prompt Template]
\refstepcounter{examplebox}\label{box:theoremqa}
\texttt{<|im\_start|>system\\
You're a helpful assistant.\\
<|im\_end|>\\
<|im\_start|>user\\
\{prompt\}\\
You are solving Math problems.\\
Turn X:\\
State:\\
<image>\\
+ (Question)\\
You have Y actions left. Always output: <think>[Your thoughts]</think><answer>[your answer]</answer> with no extra text. Strictly follow this format. Max response length: Z words (tokens).\\
<|im\_end|>\\
<|im\_start|>assistant\\
...\\
<|im\_end|>}
\end{tcolorbox}

\section{Model Evaluation Details}
\label{sec:model-detail}
We present the model used to evaluate answer repetition in Table~\ref{tab:hf-model-names}.

\begin{table}[t]
\centering
\caption{Hugging Face model names used in the unique answer ratio evaluation.}
\begin{tabular}{ll}
\toprule
\textbf{Method} & \textbf{Model (Hugging Face name)} \\
\midrule
DAPO & Qwen/Qwen2.5-32B \\
     & BytedTsinghua-SIA/DAPO-Qwen-32B \\
\midrule
Dr.\ GRPO & Qwen/Qwen2.5-Math-7B \\
          & sail/Qwen2.5-Math-7B-Oat-Zero \\
          & Qwen/Qwen2.5-Math-1.5B \\
          & sail/Qwen2.5-Math-1.5B-Oat-Zero \\
\midrule
GRPO & Qwen/Qwen2.5-0.5B \\
     & hkust-nlp/Qwen-2.5-0.5B-SimpleRL-Zoo\\
     & Qwen/Qwen2.5-1.5B \\
     & hkust-nlp/Qwen-2.5-1.5B-SimpleRL-Zoo \\
     & Qwen/Qwen2.5-7B \\
     & hkust-nlp/Qwen-2.5-7B-SimpleRL-Zoo \\
     & Qwen/Qwen2.5-14B \\
     & hkust-nlp/Qwen-2.5-14B-SimpleRL-Zoo \\
\midrule
PPO & Qwen/Qwen2.5-Math-7B \\
    & RLHFlow/Qwen2.5-7B-PPO-Zero \\
    & Qwen/Qwen2.5-3B-Instruct \\
    & LichengLiu03/Qwen2.5-3B-UFO \\
    
\bottomrule
\end{tabular}
\label{tab:hf-model-names}
\end{table}

\section{Case Analyses}

We investigate the impact of multi-turn reinforcement learning (RL) on large language models (LLMs) through a series of curated examples across distinct training stages. These case studies (shown in Boxes~\ref{box:case1}--\ref{box:case4}) highlight the evolving dynamics of exploration, convergence, and reasoning quality throughout training.

\subsection{Case 1: Pre-training Behavior}
Before any reinforcement learning, we observe the model’s default multi-turn reasoning behavior in a symbolic pattern-matching task (Box~\ref{box:case1}). The model is asked to recover a missing variable \(X\) from a repeating alphabetic value pattern, given partial information and a constraint on the sum of values in a specific word. In Turn 1, the model identifies the relevant positions in the word “numeric” and proposes an initial guess for \(X\). As feedback indicates the answer is incorrect, the model progressively refines its understanding: it attempts to align characters in the input word with their positions in the pattern and adjusts its value for \(X\).

Despite making several wrong guesses, the model demonstrates \textbf{adaptive behavior across turns}: it updates its assumptions, introduces new hypotheses, and makes meaningful structural progress (e.g., recognizing the 8-length cycle). However, it ultimately fails to reach the correct solution within the available steps. The case shows that \textbf{pretrained models already possess multi-step reflective capabilities and can utilize external feedback to revise their reasoning}, even without explicit training for multi-turn alignment. It suggest that reinforcement learning has the potential to further stabilize and guide emergent reasoning process toward convergence.

\subsection{Case 2: Post Single-turn RL}
After reinforcement learning with single-step reward feedback, the model demonstrates drastically different behavior from its pretrained counterpart (Box~\ref{box:case2}). When tasked with identifying the variable \(X\) in a cyclic pattern-based word problem, the model immediately commits to a single interpretation. In Turn 1, it attempts a symbolic derivation by aligning the letter values of “numeric” with a fixed periodic pattern, then solving \(X - 3 = -1\). However, this derivation mistakenly assumes that the letters in “numeric” correspond to the first 7 elements of the pattern without justifying the mapping. More notably, this exact sequence of logic and answer is \textbf{repeated identically} in Turns 2 through 5.

\textbf{The case reveals that single-turn RL induces brittle, overconfident behavior}: once the model settles on a trajectory during initial inference, it does not reconsider alternative hypotheses or respond meaningfully to corrective feedback. The reward optimization has led to collapse in exploration, as each turn simply replays the same incorrect reasoning with no adaptation. In contrast to the pretraining stage, where the model at least attempts different strategies, this behavior illustrates a major drawback of single-step reward supervision: it teaches the model what to say once, but not how to revise when it’s wrong.

\subsection{Case 3: Success Adaptation to Feedback Through Multi-turn RL with UFO }
This example illustrates the effectiveness of multi-turn reinforcement learning (Box~\ref{box:case3}). The model is prompted to determine the sum of all positive integers \(n\) for which \(\frac{n + 18}{n}\) is an integer. In Turn 1, it begins by simplifying the expression to \(1 + \frac{18}{n}\), and attempts a partial answer without listing all divisors. Upon receiving feedback, the model updates its understanding in Turn 2 by enumerating all positive divisors of 18. By Turn 3, it completes the reasoning process by summing those divisors correctly, arriving at the correct final answer of 39.

The case shows a successful case of multi-turn self-correction, where the model refines its reasoning incrementally in response to feedback. Each turn builds on the previous one: the model first identifies the mathematical form, then retrieves the correct domain knowledge (divisors), and finally executes a complete and valid computation. Unlike single-turn RL, \textbf{the multi-turn reward structure encourages adaptive reasoning and gradual convergence.} The example shows that properly trained multi-turn agents can both reason correctly and update their behavior mid-trajectory in pursuit of a verifiable answer.

\subsection{Case 4: Reasoning Drift of Multi-turn RL with UFO}
We examine a failure trajectory under multi-turn reinforcement learning where the model arrives at the correct final answer through incorrect or misaligned reasoning. As shown in Box~\ref{box:case4}, the task is to find the value of \(X\) such that the matrix \(\begin{pmatrix} X & 3 \\ -1 & 7 \end{pmatrix}\) has an inverse consistent with the given answer “17.” In the initial turn, the model misinterprets the problem structure by assuming that the matrix inverse itself is the scalar 17, and raises a structural objection. In the next turn, it shifts to a more plausible interpretation where 17 is assumed to be the determinant. This leads to the formulation of the correct equation \(7X + 3 = 17\), though the value of \(X\) is still miscalculated at that stage.

By the third turn, the model correctly solves for \(X = 2\), but then presents a flawed expression for the inverse matrix, misapplying the standard inversion formula for \(2 \times 2\) matrices. Despite producing the right numerical solution, the trajectory contains conceptual errors related to matrix algebra, including an incorrect cofactor layout. This case highlights a form of reasoning drift: \textbf{while the final output aligns with the ground truth, the intermediate steps do not constitute a logically valid derivation.} Such behavior reveals a potential failure of reward design in multi-turn RL, where correctness of the answer is reinforced even in the absence of faithful reasoning.

\begin{table}[t]
\centering
\caption{Comparison of multi-turn reasoning behaviors across training stages.}
\begin{tabular}{@{}llllll@{}}
\toprule
\textbf{Case} & \textbf{Stage} & \textbf{Exploration} & \textbf{Convergence} & \textbf{Reasoning Quality} & \textbf{Failure Mode} \\
\midrule
1 & Pre-RL           & High      & No    & Incomplete         & Early guessing \\
2 & Single-turn RL   & None      & No    & Repetitive         & Overfitting     \\
3 & Multi-turn RL   & Moderate  & Yes   & Correct & aligned              \\
4 & Multi-turn RL   & Moderate  & Yes   & Incorrect          & Reasoning drift \\
\bottomrule
\end{tabular}
\label{tab:cases_summary}
\end{table}

To summarize (as in Table~\ref{tab:cases_summary}), multi-turn RL with UFO improves convergence and stabilizes reasoning, but naive reward design can still lead to misalignment between thought and outcome. It highlights the need for evaluation frameworks to go beyond final answer correctness and explicitly assess reasoning integrity throughout the trajectory.

\begin{tcolorbox}[breakable, colback=gray!2, colframe=black, boxrule=0.5pt, arc=2pt, width=\linewidth, title=Box 4: Multi-Turn Rollout before training]
\refstepcounter{examplebox}\label{box:case1}
\textbf{Turn 1:} \\
\textbf{State:}
The letters of the alphabet are given numeric values based on the two conditions below.

$\bullet$  Only the numeric values of $-2,$ $-1,$ $0,$ $1$ and $2$ are used.

$\bullet$  Starting with A and going through Z, a numeric value is assigned to each letter according to the following pattern: $$
1, 2, X, 0, -1, -2, -1, 0, 1, 2, 1, 0, -1, -2, -1, 0,\ldots
$$

Two complete cycles of the pattern are shown above. The letter A has a value of $1,$ B has a value of $2,$ F has a value of $-2$ and Z has a value of $2$. The sum of the numeric values of the letters in the word ``numeric" is -1. What is the value of unknown variable X?
You have 10 actions left. Always output: \texttt{<think>} [Your thoughts] \texttt{<think>} \texttt{<answer>} [your answer] \texttt{<answer>} with no extra text. Strictly follow this format. Max response length: 100 words (tokens). \\
\textbf{Output:} \\
\texttt{<think>}First, let's identify the positions of each letter in the word ``numeric" based on the pattern provided. We know the values of A, B, F, and Z. We need to determine the value of X and then find the values of N, M, E, and C to sum up to -1.\texttt{</think>} \\
\texttt{<answer>}X = 2\texttt{</answer>} \\
Reward: 0.0

\vspace{4pt}
\textbf{Turn 2:} \\
\textbf{State:}
Incorrect. Please try agin.
You have 3 actions left. Always output: \texttt{<think>} [Your thoughts] \texttt{<think>} \texttt{<answer>} [your answer] \texttt{<answer>} with no extra text. Strictly follow this format. Max response length: 100 words (tokens). \\
\textbf{Output:} \\
\texttt{<think>}Given the incorrect answer, let's re-examine the pattern and the values assigned to each letter. The word ``numeric" has the letters N, U, M, B, E, I, C. From the provided pattern, we know the values of A, B, F, and Z. To find the value of X, we need to identify the value for N and then determine the pattern for subsequent letters.\texttt{</think>} \\
\texttt{<answer>}X = 0\texttt{</answer>} \\
Reward: 0.0

\vspace{4pt}
\textbf{Turn 3:} \\
\textbf{State:}
Incorrect. Please try agin.
You have 2 actions left. Always output: \texttt{<think>} [Your thoughts] \texttt{<think>} \texttt{<answer>} [your answer] \texttt{<answer>} with no extra text. Strictly follow this format. Max response length: 100 words (tokens). \\
\textbf{Output:} \\
\texttt{<think>}Given the incorrect answer, let's re-evaluate the pattern and the values assigned to each letter in the word ``numeric". The values for A, B, F, and Z are known. The pattern seems to repeat every 8 letters, and the values cycle through 1, 2, X, 0, -1, -2, -1, 0. We need to determine the value of X such that the sum of the values of the letters in ``numeric" equals -1.\texttt{</think>} \\
\texttt{<answer>}X = -1\texttt{</answer>} \\
Reward: 0.0

\vspace{4pt}
\textbf{Turn 4:} \\
\textbf{State:}
Incorrect. Please try agin.
You have 1 actions left. Always output: \texttt{<think>} [Your thoughts] \texttt{<think>} \texttt{<answer>} [your answer] \texttt{<answer>} with no extra text. Strictly follow this format. Max response length: 100 words (tokens). \\
\textbf{Output:} \\
\texttt{<think>}Given the incorrect answer, let's re-examine the pattern and the values assigned to each letter in the word ``numeric". The values for A, B, F, and Z are known, and the pattern seems to repeat every 8 letters with values 1, 2, X, 0, -1, -2, -1, 0. To find the value of X such that the sum of the values of the letters in ``numeric" equals -1, we need to assign values to N, U, M, B, E, I, C and ensure their sum is -1\texttt{</think>} \\
\texttt{<answer>}X = 1\texttt{</answer>} \\
Reward: 0.0 \\

\vspace{4pt}
\textbf{Turn 5:} \\
\textbf{State:}
Incorrect. Please try agin.
You have 0 actions left. Always output: \texttt{<think>} [Your thoughts] \texttt{<think>} \texttt{<answer>} [your answer] \texttt{<answer>} with no extra text. Strictly follow this format. Max response length: 100 words (tokens). \\
\textbf{Output:} \\
\texttt{<think>}Given the incorrect answer, let's re-examine the pattern and the values assigned to each letter in the word ``numeric". The values for A, B, F, and Z are known, and the pattern seems to repeat every 8 letters with values 1, 2, X, 0, -1, -2, -1, 0. To find the value of X such that the sum of the values of the letters in ``numeric" equals -1, we need to assign values to N, U, M, B, E, I, C and ensure their sum is -1.\texttt{</think>} \\
\texttt{<answer>}X = -1\texttt{</answer>} \\
Reward: 0.0
\end{tcolorbox}

\begin{tcolorbox}[breakable, colback=gray!2, colframe=black, boxrule=0.5pt, arc=2pt, width=\linewidth, title=Box 5: Model rollout after single-turn RL training]
\refstepcounter{examplebox}\label{box:case2}
\textbf{Turn 1:} \\
\textbf{State:}
The letters of the alphabet are given numeric values based on the two conditions below.

$\bullet$  Only the numeric values of $-2,$ $-1,$ $0,$ $1$ and $2$ are used.

$\bullet$  Starting with A and going through Z, a numeric value is assigned to each letter according to the following pattern: $$
1, 2, X, 0, -1, -2, -1, 0, 1, 2, 1, 0, -1, -2, -1, 0,\ldots
$$

Two complete cycles of the pattern are shown above. The letter A has a value of $1,$ B has a value of $2,$ F has a value of $-2$ and Z has a value of $2$. The sum of the numeric values of the letters in the word ``numeric" is -1. What is the value of unknown variable X?
You have 10 actions left. Always output: \texttt{<think>} [Your thoughts] \texttt{<think>} \texttt{<answer>} [your answer] \texttt{<answer>} with no extra text. Strictly follow this format. Max response length: 100 words (tokens). \\
\textbf{Output:} \\
\texttt{<think>}For the word ``numeric", the values of the letters are as follows: N has a value of 1, U has a value of 2, M has a value of 0, E has a value of -1, R has a value of -2, I has a value of -1, C has a value of 1. The sum of these values is \(1 + 2 + 0 - 1 - 2 - 1 + 1 = 0\), which contradicts the given sum of -1. However, we can re-evaluate the pattern and the values. The pattern repeats every 16 letters, and the values repeat as well. The word ``numeric" consists of 7 letters, so we need to find the sum of the first 7 values in the pattern: \(1, 2, X, 0, -1, -2, -1\). We know the sum of these values is -1. So, we have \(1 + 2 + X + 0 - 1 - 2 - 1 = -1\), which simplifies to \(X - 3 = -1\), and thus \(X = 2\).\texttt{</think>} \\
\texttt{<answer>}2\texttt{</answer>} \\
Reward: 0.0

\vspace{4pt}
\textbf{Turn 2:} \\
\textbf{State:}
Incorrect. Please try agin.
You have 3 actions left. Always output: \texttt{<think>} [Your thoughts] \texttt{<think>} \texttt{<answer>} [your answer] \texttt{<answer>} with no extra text. Strictly follow this format. Max response length: 100 words (tokens). \\
\textbf{Output:} \\
\texttt{<think>}For the word ``numeric", the values of the letters are as follows: N has a value of 1, U has a value of 2, M has a value of 0, E has a value of -1, R has a value of -2, I has a value of -1, C has a value of 1. The sum of these values is \(1 + 2 + 0 - 1 - 2 - 1 + 1 = 0\), which contradicts the given sum of -1. The pattern repeats every 16 letters, and the values repeat as \(1, 2, X, 0, -1, -2, -1, 0, 1, 2, 1, 0, -1, -2, -1, 0\). For the word ``numeric", the values are \(1, 2, X, 0, -1, -2, -1\). The sum of these values is \(1 + 2 + X + 0 - 1 - 2 - 1 = -1\), which simplifies to \(X - 3 = -1\), and thus \(X = 2\).\texttt{</think>} \\
\texttt{<answer>}2\texttt{</answer>} \\
Reward: 0.0

\vspace{4pt}
\textbf{Turn 3:} \\
\textbf{State:}
Incorrect. Please try agin.
You have 2 actions left. Always output: \texttt{<think>} [Your thoughts] \texttt{<think>} \texttt{<answer>} [your answer] \texttt{<answer>} with no extra text. Strictly follow this format. Max response length: 100 words (tokens). \\
\textbf{Output:} \\
\texttt{<think>}For the word ``numeric", the values of the letters are as follows: N has a value of 1, U has a value of 2, M has a value of 0, E has a value of -1, R has a value of -2, I has a value of -1, C has a value of 1. The sum of these values is \(1 + 2 + 0 - 1 - 2 - 1 + 1 = 0\), which contradicts the given sum of -1. The pattern repeats every 16 letters, and the values repeat as \(1, 2, X, 0, -1, -2, -1, 0, 1, 2, 1, 0, -1, -2, -1, 0\). For the word ``numeric", the values are \(1, 2, X, 0, -1, -2, -1\). The sum of these values is \(1 + 2 + X + 0 - 1 - 2 - 1 = -1\), which simplifies to \(X - 3 = -1\), and thus \(X = 2\).\texttt{</think>} \\
\texttt{<answer>}2\texttt{</answer>} \\
Reward: 0.0

\vspace{4pt}
\textbf{Turn 4:} \\
\textbf{State:}
Incorrect. Please try agin.
You have 1 actions left. Always output: \texttt{<think>} [Your thoughts] \texttt{<think>} \texttt{<answer>} [your answer] \texttt{<answer>} with no extra text. Strictly follow this format. Max response length: 100 words (tokens). \\
\textbf{Output:} \\
\texttt{<think>}For the word ``numeric", the values of the letters are as follows: N has a value of 1, U has a value of 2, M has a value of 0, E has a value of -1, R has a value of -2, I has a value of -1, C has a value of 1. The sum of these values is \(1 + 2 + 0 - 1 - 2 - 1 + 1 = 0\), which contradicts the given sum of -1. The pattern repeats every 16 letters, and the values repeat as \(1, 2, X, 0, -1, -2, -1, 0, 1, 2, 1, 0, -1, -2, -1, 0\). For the word ``numeric", the values are \(1, 2, X, 0, -1, -2, -1\). The sum of these values is \(1 + 2 + X + 0 - 1 - 2 - 1 = -1\), which simplifies to \(X - 3 = -1\), and thus \(X = 2\).\texttt{</think>} \\
\texttt{<answer>}2\texttt{</answer>} \\
Reward: 0.0 \\
\vspace{4pt}
\textbf{Turn 5:} \\
\textbf{State:}
Incorrect. Please try agin.
You have 0 actions left. Always output: \texttt{<think>} [Your thoughts] \texttt{<think>} \texttt{<answer>} [your answer] \texttt{<answer>} with no extra text. Strictly follow this format. Max response length: 100 words (tokens). \\
\textbf{Output:} \\
\texttt{<think>}For the word ``numeric", the values of the letters are as follows: N has a value of 1, U has a value of 2, M has a value of 0, E has a value of -1, R has a value of -2, I has a value of -1, C has a value of 1. The sum of these values is \(1 + 2 + 0 - 1 - 2 - 1 + 1 = 0\), which contradicts the given sum of -1. The pattern repeats every 16 letters, and the values repeat as \(1, 2, X, 0, -1, -2, -1, 0, 1, 2, 1, 0, -1, -2, -1, 0\). For the word ``numeric", the values are \(1, 2, X, 0, -1, -2, -1\). The sum of these values is \(1 + 2 + X + 0 - 1 - 2 - 1 = -1\), which simplifies to \(X - 3 = -1\), and thus \(X = 2\).\texttt{</think>} \\
\texttt{<answer>}2\texttt{</answer>} \\
Reward: 0.0
\end{tcolorbox}

\begin{tcolorbox}[breakable, colback=gray!2, colframe=black, boxrule=0.5pt, arc=2pt, width=\linewidth, title=Box 6: Model rollout after multi-turn RL training with UFO (Success Case)]
\refstepcounter{examplebox}\label{box:case3}
\textbf{Turn 1:} \\
\textbf{State:} \\
Determine the sum of all positive integer values of \(n\) such that \(\frac{n+18}{n}\) is an integer. \\
You have 10 actions left. Always output: \texttt{<think>} [Your thoughts] \texttt{</think>} \texttt{<answer>} [your answer] \texttt{</answer>} with no extra text. Strictly follow this format. Max response length: 100 words. \\
\textbf{Output:} \\
\texttt{<think>}First, I need to understand when \(\frac{n+18}{n}\) is an integer. This fraction can be simplified to \(1 + \frac{18}{n}\). For this to be an integer, \(n\) must be a divisor of 18.\texttt{</think>} \\
\texttt{<answer>}18 and 1 (since \(1 + \frac{18}{18} = 2\) and \(1 + \frac{18}{1} = 19\) are integers)\texttt{</answer>} \\
Reward: 0.0

\vspace{4pt}
\textbf{Turn 2:} \\
\textbf{State:} \\
Incorrect. Please think again. \\
You have 9 actions left. \\
\textbf{Output:} \\
\texttt{<think>}Given \(\frac{n+18}{n}\) is an integer, we have \(1 + \frac{18}{n}\) as an integer. This means \(n\) must be a divisor of 18. The positive divisors of 18 are 1, 2, 3, 6, 9, and 18. I need to verify which of these values make the expression an integer.\texttt{</think>} \\
\texttt{<answer>}1, 2, 3, 6, 9, 18\texttt{</answer>} \\
Reward: 0.0

\vspace{4pt}
\textbf{Turn 3:} \\
\textbf{State:} \\
Incorrect. Please think again. \\
You have 8 actions left. \\
\textbf{Output:} \\
\texttt{<think>}Upon reviewing, I realize the divisors of 18 are 1, 2, 3, 6, 9, and 18. For each of these values, \(1 + \frac{18}{n}\) is indeed an integer. Therefore, the sum of all positive integer values of \(n\) is \(1 + 2 + 3 + 6 + 9 + 18 = 39\).\texttt{</think>} \\
\texttt{<answer>}39\texttt{</answer>} \\
Reward: 0.25
\end{tcolorbox}

\begin{tcolorbox}[breakable, colback=gray!2, colframe=black, boxrule=0.5pt, arc=2pt, width=\linewidth, title=Box 7: Model rollout after multi-turn RL training with UFO (Failure Case)]
\refstepcounter{examplebox}\label{box:case4}
\textbf{Turn 1:} \\
\textbf{State:} \\
Find the inverse of the matrix
\[
\begin{pmatrix} X & 3 \\ -1 & 7 \end{pmatrix}.
\]
If the inverse does not exist, then enter the zero matrix. The answer is 17. What is the value of unknown variable \(X\)? \\
You have 10 actions left. Always output: \texttt{<think>} [Your thoughts] \texttt{</think>} \texttt{<answer>} [your answer] \texttt{</answer>} with no extra text. Strictly follow this format. Max response length: 100 words. \\
\textbf{Output:} \\
\texttt{<think>}Since the inverse of a matrix is given as 17, and the matrix provided is 2x2, the inverse should be a 2x2 matrix. However, the given inverse of 17 does not align with the structure of a 2x2 matrix. The problem might have a mistake in the provided inverse value or the matrix structure.\texttt{</think>} \\
\texttt{<answer>}There is an inconsistency in the problem statement. The given inverse of 17 does not match the structure of a 2x2 matrix inverse.\texttt{</answer>} \\
Reward: 0.0

\vspace{4pt}
\textbf{Turn 2:} \\
\textbf{State:} \\
Incorrect. Please think again. \\
You have 9 actions left. \\
\textbf{Output:} \\
\texttt{<think>}Given the inverse of the matrix is provided as 17, which does not fit the expected format of a 2x2 matrix inverse (which should be a 2x2 matrix), it seems there might be a misunderstanding. However, since the problem statement specifies the inverse as 17, and considering the structure of the matrix, it implies the inverse is a scalar value. For a 2x2 matrix \(\begin{pmatrix} a & b \\ c & d \end{pmatrix}\), the inverse is \(\frac{1}{ad-bc}\begin{pmatrix} d & -b \\ -c & a \end{pmatrix}\). If the inverse is simply 17, it suggests \(ad - bc = 17\). Given the matrix \(\begin{pmatrix} X & 3 \\ -1 & 7 \end{pmatrix}\), \(ad - bc = X \cdot 7 - 3 \cdot (-1) = 7X + 3\). Thus, \(7X + 3 = 17\).\texttt{</think>} \\
\texttt{<answer>}7\texttt{</answer>} \\
Reward: 0.0

\vspace{4pt}
\textbf{Turn 3:} \\
\textbf{State:} \\
Incorrect. Please think again. \\
You have 8 actions left. \\
\textbf{Output:} \\
\texttt{<think>}Given the matrix \(\begin{pmatrix} X & 3 \\ -1 & 7 \end{pmatrix}\), the condition for the inverse to be a scalar 17 is \(7X + 3 = 17\). Solving for \(X\) gives \(7X = 14\), thus \(X = 2\). The inverse of the matrix should be \(\frac{1}{17}\begin{pmatrix} 7 & -3 \\ 1 & X \end{pmatrix}\), which simplifies to \(\begin{pmatrix} \frac{7}{17} & -\frac{3}{17} \\ \frac{1}{17} & 2 \end{pmatrix}\), not just 17.\texttt{</think>} \\
\texttt{<answer>}2\texttt{</answer>} \\
Reward: 0.25
\end{tcolorbox}